\PassOptionsToPackage{table,xcdraw}{xcolor}
\documentclass[sigconf,review=false]{acmart}

\usepackage{graphicx}

\usepackage{bm}
\usepackage{booktabs}
\usepackage[table]{xcolor}
\usepackage{amsmath}  
\definecolor{green}{RGB}{0,128,0}  
\definecolor{red}{RGB}{255,0,0}    
\usepackage{multirow}
\usepackage{amsmath}
\usepackage{amssymb}
\usepackage{array}
\usepackage{longtable}
\usepackage{tabularx}
\usepackage{supertabular}
\usepackage{makecell}
\usepackage{array}
\usepackage{hyperref}
\usepackage[capitalize,noabbrev]{cleveref}
\usepackage{enumitem}

\usepackage{microtype}
\usepackage{graphicx}
\usepackage{subfigure}
\usepackage{booktabs} 

\usepackage{mathtools}
\usepackage{amsthm}
\usepackage{titlesec}
\usepackage{graphicx}
\usepackage{subcaption}
\usepackage{caption}
\usepackage{float} 
\usepackage{stfloats}

\titleformat{\paragraph}[runin] {\bfseries\normalfont}{\theparagraph}{1em}{}

\renewcommand\footnotetextcopyrightpermission[1]{}

\AtBeginDocument{%
  }

\setcopyright{acmlicensed}
\copyrightyear{2025}
\acmYear{2025}
\acmDOI{XXXXXXX.XXXXXXX}
\acmConference[Anonymous]{Make sure to enter the correct conference title from your rights confirmation email}{2025}{}
\acmISBN{978-1-4503-XXXX-X/2018/06}

\usepackage{listings}
\usepackage{color}

\definecolor{dkgreen}{rgb}{0,0.6,0}
\definecolor{gray}{rgb}{0.5,0.5,0.5}
\definecolor{mauve}{rgb}{0.58,0,0.82}

\begin{document}

\title{FedMABench: Benchmarking Mobile Agents on Decentralized Heterogeneous User Data}


\author{Wenhao Wang\textsuperscript{*}}
 \affiliation{
   \institution{Zhejiang University}
   \institution{Shanghai AI Laboratory}
   \city{Hangzhou}
   \country{China}
 }
\email{12321254@zju.edu.cn}

\author{Zijie Yu\textsuperscript{*}}
 \affiliation{
   \institution{Shanghai Jiao Tong University}
   \city{Shanghai}
   \country{China}
 }
\email{zijie.yu@sjtu.edu.cn}

\author{Rui Ye}
 \affiliation{
   \institution{Shanghai Jiao Tong University}
   \city{Shanghai}
   \country{China}
 }
\email{yr991129@sjtu.edu.cn}

\author{Jianqing Zhang}
 \affiliation{
   \institution{Shanghai Jiao Tong University}
   \city{Shanghai}
   \country{China}
 }
\email{tsingz@sjtu.edu.cn}

\author{Siheng Chen\textsuperscript{†}}
 \affiliation{
   \institution{Shanghai Jiao Tong University}
   \institution{Shanghai AI Laboratory}
   \city{Shanghai}
   \country{China}
 }
\email{sihengc@sjtu.edu.cn}

\author{Yanfeng Wang\textsuperscript{†}}
 \affiliation{
    \institution{Shanghai AI Laboratory}
   \institution{Shanghai Jiao Tong University}
   \city{Shanghai}
   \country{China}
 }
\email{wangyanfeng@sjtu.edu.cn}

\renewcommand{\shortauthors}{Wenhao Wang, Zijie Yu et al.}

\begin{abstract}
Mobile agents have attracted tremendous research participation recently.
Traditional approaches to mobile agent training rely on centralized data collection, leading to high cost and limited scalability. 
Distributed training utilizing federated learning offers an alternative by harnessing real-world user data, providing scalability and reducing costs. 
However, pivotal challenges, including the absence of standardized benchmarks, hinder progress in this field.

To tackle the challenges, we introduce FedMABench, the first benchmark for federated training and evaluation of mobile agents, specifically designed for heterogeneous scenarios. FedMABench features 6 datasets with 30+ subsets, 8 federated algorithms, 10+ base models, and over 800 apps across 5 categories, providing a comprehensive framework for evaluating mobile agents across diverse environments. 
Through extensive experiments, we uncover several key insights: federated algorithms consistently outperform local training; the distribution of specific apps plays a crucial role in heterogeneity; and, even apps from distinct categories can exhibit correlations during training.
FedMABench is publicly available at: \href{https://github.com/wwh0411/FedMABench}{https://github.com/wwh0411/FedMABench} with the datasets at: \href{https://huggingface.co/datasets/wwh0411/FedMABench}{https://huggingface.co/datasets/wwh0411/FedMABench}.

\end{abstract}


\settopmatter{printacmref=false} 



\maketitle

\begingroup
\renewcommand\thefootnote{$*$}
\footnotetext{These authors contributed equally to this work.}
\renewcommand\thefootnote{$\dag$}
\footnotetext{Corresponding authors.}
\endgroup



\begin{figure*}[t]
\centering
\includegraphics[width=0.75\textwidth]{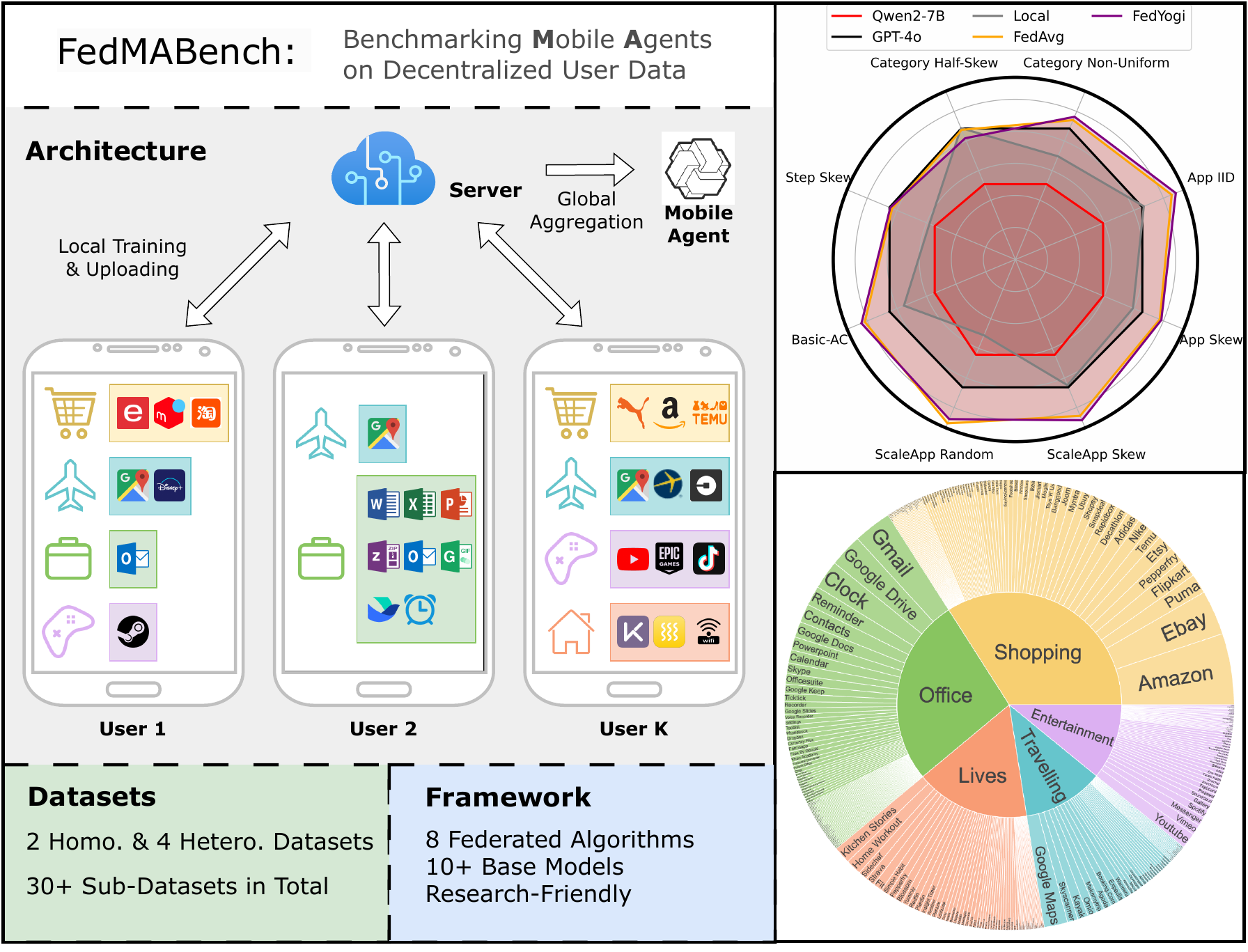}
\vspace{-2mm}
\caption{Overview of FedMABench. FedMABench is tailored for benchmarking federated mobile agents trained on distributed mobile user data with diverse types of heterogeneity. To achieve this, we construct 2 homogeneous dataset and 4 heterogeneous datasets with 30+ subsets. We also build a research-friendly framework, which integrates 8 representative federated algorithms and supports evaluation on more than 10 base models. 
Our datasets covers 877 apps across 5 categories (bottom right) and the experiments (upper right) showcase that (1) federated mobile agents achieve promising results, surpassing GPT-4o by a large margin; (2) our datasets can reveal the performance differences of mobile agents on different distributions. }
\label{fig:main}
\vspace{-2mm}
\end{figure*}

\section*{Keywords}
Mobile Agent, Federated Learning, Data Heterogeneity
\section{Introduction}
Recent advances in Vision-Language Models (VLMs) \cite{wang2021minivlmsmallerfastervisionlanguage,jin2021good, zhou2022learning} have significantly propelled the evolution of Graphical User Interface (GUI) agents \cite{baiDigiRLTrainingInTheWild2024, wangDistRLAsynchronousDistributed2024, wangMobileAgentBenchEfficientUserFriendly2024}. 
GUI agents on mobile phones, known as \textbf{Mobile Agents}, are capable of automating complex tasks, thereby significantly reducing human workload. 
Mobile agents have demonstrated promising potential across a wide range of applications \cite{liuAutoGLMAutonomousFoundation2024}.

The traditional approach for mobile agents largely depends on centralized data collection and training \cite{hongCogAgentVisualLanguage2023, dorkaTrainingVisionLanguage2024, chen2024octopusv2ondevicelanguage}, which, although effective, leads to several challenges such as high costs and limited scalability \cite{sunOSGenesisAutomatingGUI2024}. 
Meanwhile, the frequent use of mobile phones by users worldwide naturally generates valuable supervisory information, which can serve as a rich data source for training mobile agents. However, this wealth of high-quality data remains underutilized, as it cannot be publicly shared due to privacy concerns \cite{xiong2025pilotbuildingfederatedmultimodal}.
Therefore, data from real-world mobile users must be utilized in a distributed manner, where each client locally collects and trains on its own data without direct data transmission.

Continuing to improve the quality and coverage of mobile agents necessitates the development of distributed data collection and training \cite{wang2025fedmobileagenttrainingmobileagents}.
Distributed training mobile agents on user data offers two key advantages: 
(1) In consideration of the billions of phone users worldwide, collecting data directly from real-world users enables unprecedented scalability.
(2) The data collection and annotation costs can be significantly reduced, as user data is an incidental by-product of daily phone usage. 
Additionally, privacy concerns surrounding the collection of personal data can be effectively mitigated through the application of Federated Learning (FL) \cite{fedavg, federatedscopellm, wang2024florafederatedfinetuninglarge}, which ensures that sensitive information remains decentralized, thus fostering greater user trust and ensuring compliance with privacy regulations.



Despite the promising potential of training mobile agents on distributed user data through FL, a critical challenge persists: \textbf{the absence of standardized benchmarks for federated mobile agents}, which impedes comparisons and advancements in this field.
In this context, (1) without diverse and heterogeneous datasets, research efforts cannot effectively address the issue of heterogeneity, which is crucial to utilizing distributed phone usage trajectories.
(2) Without an efficient and unified framework, future research may give rise to varied training and evaluation protocols, complicating re-implementations and heightening the risk of unfair comparisons.




To address these challenges, we introduce \textbf{FedMABench}, the first benchmark specifically designed for federated training and evaluation of mobile agents, with three key features:
(1) \textbf{Comprehensiveness}: FedMABench provides a comprehensive framework that integrates eight federated algorithms and supports over ten base models. The evaluation metrics include two performance indicators for both high-level and low-level training, establishing a solid foundation for future research and development.
(2) \textbf{Diversity}: 
FedMABench includes thousands of tasks, spanning over 800 apps across five categories from two distinct data sources, yielding substantial diversity.
(3) \textbf{Heterogeneity}:
FedMABench puts strong emphasize on heterogeneous scenarios to promote further research.
We incorporate 30+ datasets derived from the original Android Control and Android in the Wild datasets \cite{rawlesAndroidWildLargeScale2023, liEffectsDataScale2024}, carefully curated to ensure fair and standardized training and evaluation setups. 

Specifically, our datasets address three typical types of heterogeneity, reflecting the diverse mobile usage patterns and preferences of users worldwide:
(1) \textbf{App Category Distribution}:  
Each app category addresses a specific type of user need. Since mobile phone usage varies based on users' different needs, the distribution of app categories becomes inherently heterogeneous.
We categorize all apps into five groups and create 6 sub-datasets reflecting different category distributions.
(2) \textbf{Specific App Preference}: Users exhibit varying preferences for specific apps even with the same function. We construct two series of datasets: one focusing on underlying the differences between apps by selecting the top five apps for experiments and the other expanding the scope with more clients and apps for further validation.
(3) \textbf{Two-Level Sample Counts}: 
Mobile agent datasets comprise different number of episodes, where differences in users' tasks and usage patterns lead to additional variations in the number of steps required to complete each episode. This two-level structure, manifesting in both episode and step counts per client, results in noticeable heterogeneity across clients.
%
\begin{table*}[t]
\centering
\setlength\tabcolsep{6pt}
\caption{Summary of the six dataset series in FedMABench. 
N. denotes "the number of". The training set and evaluation set are combined by "+". 
Our datasets span a broad spectrum of \colorbox{red!10}{homogeneity} and \colorbox{blue!10}{heterogeneity}, encompassing a variety of apps across five categories. For Basic-AC and Basic-AitW, we report the maximum numbers of episode and step.
}
\vspace{-2mm}
\begin{tabular}{l|cccccc}
\toprule
\textbf{Dataset Name} & \textbf{Distribution Characteristic} & \textbf{N. Subsets} & \textbf{N. Clients} & \textbf{N. Apps} & \textbf{N. Episodes} & \textbf{N. Steps}  \\ 
\midrule
\rowcolor{red!10} Basic-AC & Homogeneous & 14 & 10-70 & 877 & 7,000+700 & 47055+4648  \\ 
\rowcolor{red!10} Basic-AitW & Homogeneous & 5 & 10-50 & - & 5,000+500 & 39394+4447  \\ 
\rowcolor{blue!10} Step-Episode & Two-Level Sample Counts & 4 & 10 & 293 & 1,000+100 & 6685+635  \\ 
\rowcolor{blue!10} Category-Level & App Category Distribution & 6 & 5 & 52 & 1,000+100 & 7127+703  \\ 
\rowcolor{blue!10} App-Level & Specific App Preference & 4 & 5 & 5 & 750+100 & 4456+574 \\ 
\rowcolor{blue!10} ScaleApp & Specific App Preference & 3 & 30 & 30 & 2,500+250 & 15700+1691  \\ 
\bottomrule
\end{tabular}
\label{tab:summay_datasets}
\vspace{-2mm}
\end{table*}


Based on FedMABench, we conduct an exhaustive empirical study to explore federated mobile agents in diverse scenarios, offering new insights into this area. 
Through extensive experiments, we make several key observations:
(1) FL algorithms consistently outperform local training, providing strong motivation for users to collaborate in exchange for more capable mobile agents.
(2) The existence of aforementioned heterogeneity negatively affects the performance of federated mobile agents.
(3) The distribution of specific apps is more fundamental to represent heterogeneity than app categories.
(4) Even apps from distinct categories can exhibit correlations during training.
In summary, our contributions are:
\begin{enumerate}[nosep, left=0.5em]
    \item We propose FedMABench, the first benchmark for federated training and evaluation of mobile agents, which is both research-friendly and comprehensive, integrating eight federated algorithms and supporting 10+ base models.
    \item We release 6 datasets with 30+ subsets, specifically targeted at three typical types of heterogeneity across various scenarios, simulating real-world user behavior on diverse apps.
    \item We conduct extensive experiments to thoroughly investigate the training of federated mobile agents on distributed data with diverse distributions, revealing insightful discoveries.
\end{enumerate}

\section{Related Work}
\label{sec:related_work}


\subsection{Current Mobile Agents on Centralized Data}
The emergence of VLMs \cite{vlmsurver} has revolutionized phone automation by facilitating more adaptive, contextually aware, and intelligent interactions with mobile devices \cite{mobilesurvey}. 
The evolution of mobile agents has undergone several pivotal advancements, with modern models exhibiting enhanced capabilities in processing multi-modal information, discerning user intentions, and autonomously performing intricate user tasks \cite{zhangUIHawkUnleashingScreen2024, nongMobileFlowMultimodalLLM2024}.




\textbf{Datasets.}
Acquiring training trajectories for mobile agents presents considerable challenges. 
The research community has invested tremendous efforts into constructing high-quality datasets for mobile agents \cite{rawlesAndroidWildLargeScale2023, zhou2024webarenarealisticwebenvironment, zhangAndroidZooChainofActionThought2024}. 
However, existing approaches primarily rely on manual curation, rendering data collection both costly and inefficient, and limiting scalability
\cite{gao2024mobileviewslargescalemobilegui, liFerretUI2Mastering2024}. 

\textbf{Benchmarks.}
Several works have sought to establish efficient benchmarks for mobile agents \cite{zhangAgentOhanaDesignUnified2024, wangMobileAgentBenchEfficientUserFriendly2024, rawlesAndroidWorldDynamicBenchmarking2024}. Yet, none of them is tailored for the distributed or federated training of mobile agents. 
While there are benchmarks for the federated training of Large Language Models (LLMs) \cite{ye2024fedllmbenchrealisticbenchmarksfederated, ye2024openfedllm, wu2024fedbiot}, they are not applicable to mobile agent training.
This gap significantly obstructs the advancement of federated mobile agents, which offer superior scalability.

\subsection{Initial Attempts of Mobile Agents on Distributed Data}
To enhance scalability and reduce the cost of traditional mobile agents on centralized data, training on distributed user data is proposed as an alternative. 
Given the privacy sensitivity, data from mobile users cannot be transmitted or consolidated, and must remain on the local devices in a "distributed" manner. In this approach, each user trains a mobile agent locally, and the resulting models are then uploaded to a central server for aggregation, producing a global mobile agent with improved capabilities.

\textbf{Federated Mobile Agent.}
FedMobileAgent \cite{wang2025fedmobileagenttrainingmobileagents} stands as a pioneering approach that proposes distributed training for mobile agents using self-sourced data from diverse users. It leverages locally deployed VLMs to automatically annotate user instructions and integrates federated learning to collaboratively optimize a global mobile agent. 
In FedMobileAgent, the authors introduce a novel form of heterogeneity, specifically the two-level distribution: one at the data step level within a trajectory, and another at the trajectory level within the local dataset. 
However, the study falls short of further investigating the complexities of heterogeneity, 
or other real-world scenarios of diverse user phone usage.

\textbf{Challenges.}
Federated mobile agents face two major challenges:
(1) To facilitate collaboration among a large and diverse set of users with varying usage habits, it is essential to address the issue of heterogeneity \cite{ye2023heterogeneous,qu2022rethinking}. 
This heterogeneity manifests in various forms, such as differing app usage patterns, individual needs, and app preferences for similar functionalities.
However, these facets of heterogeneity remain largely unexplored, with vast potential yet to be uncovered.
(2) Currently, no publicly available datasets or benchmarks exist for training federated mobile agents. 
And it is non-trivial to effectively capture the heterogeneity that is representative of real-world scenarios by directly down-sampling from existing datasets.
In this context, our FedMABench stands out as the first comprehensive benchmark in the literature, addressing these gaps.

\begin{figure*}[t]
	\centering
	\begin{minipage}{0.24\linewidth}
		\centerline{\includegraphics[width=\textwidth]{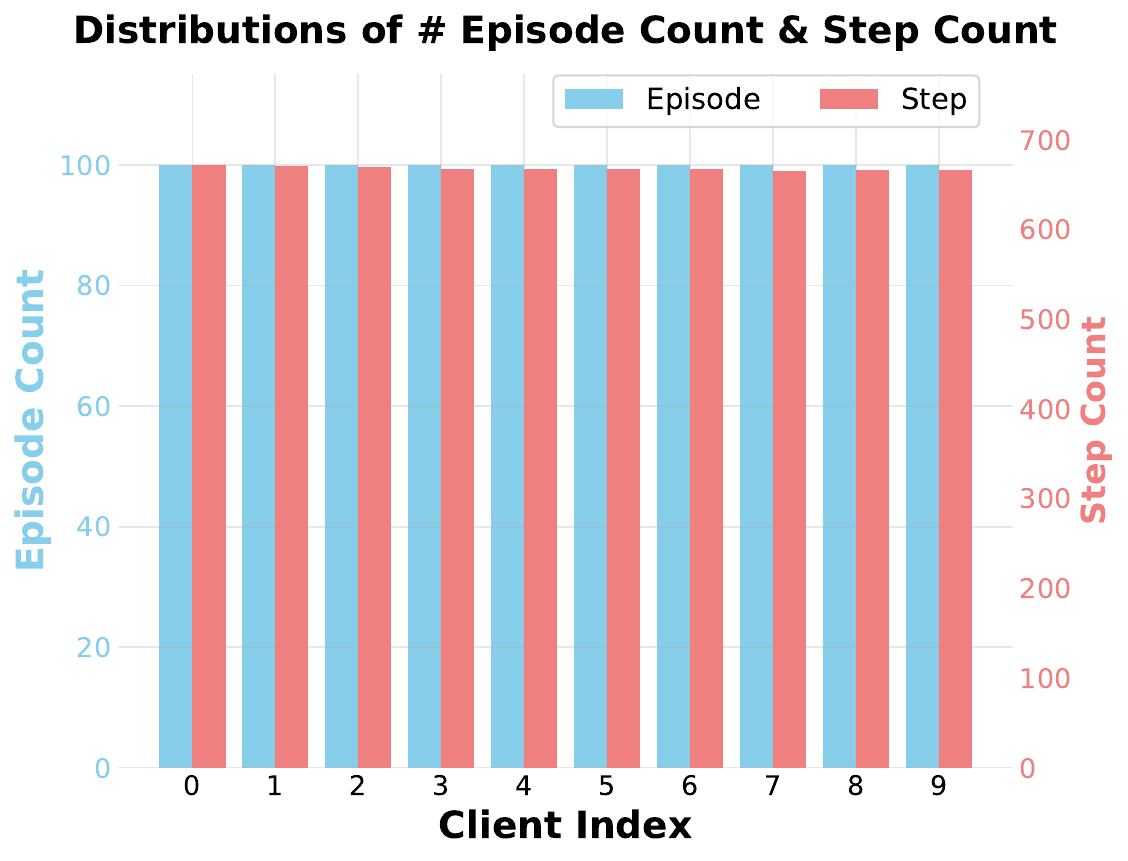}}
		\centerline{(a) Step-Episode IID}
	\end{minipage}
    \hspace{0.005\linewidth} 
    \begin{minipage}{0.24\linewidth}
		\centerline{\includegraphics[width=\textwidth]{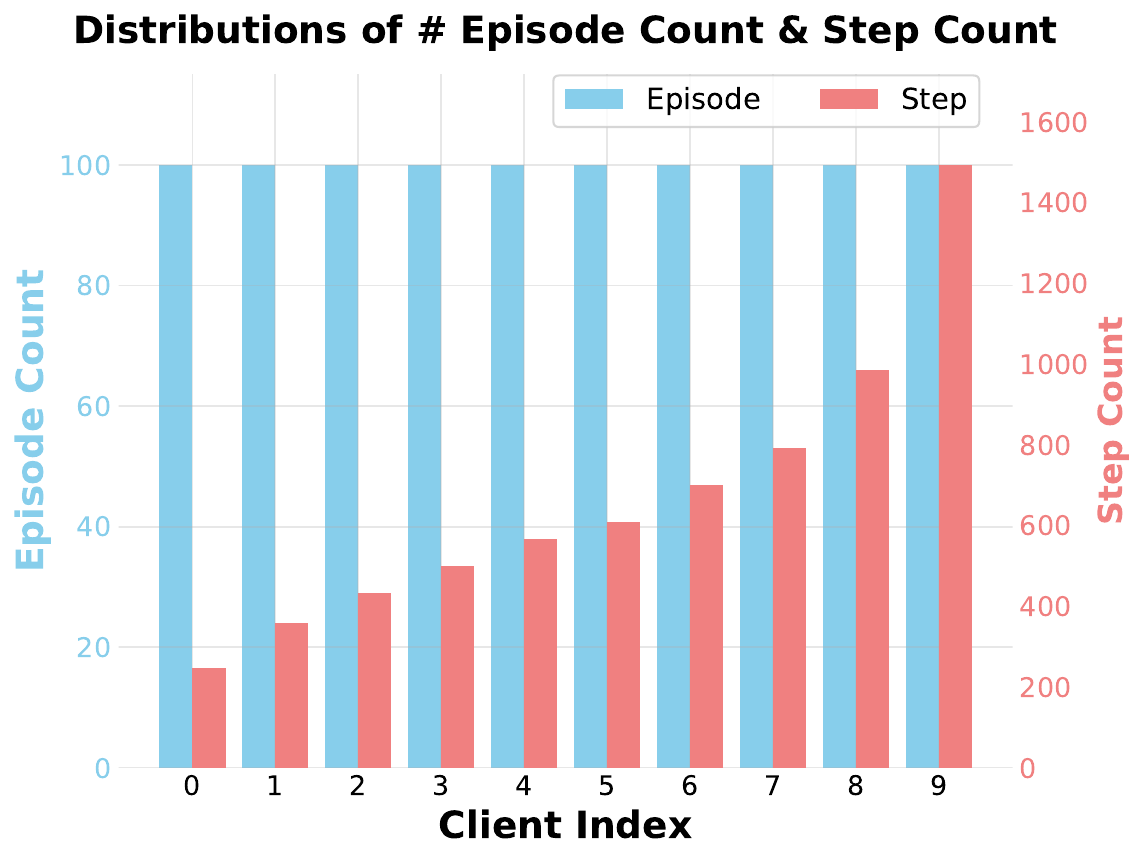}}
		\centerline{(b) Step Skew}
	\end{minipage}
    \hspace{0.005\linewidth} 
	\begin{minipage}{0.24\linewidth}
		\centerline{\includegraphics[width=\textwidth]{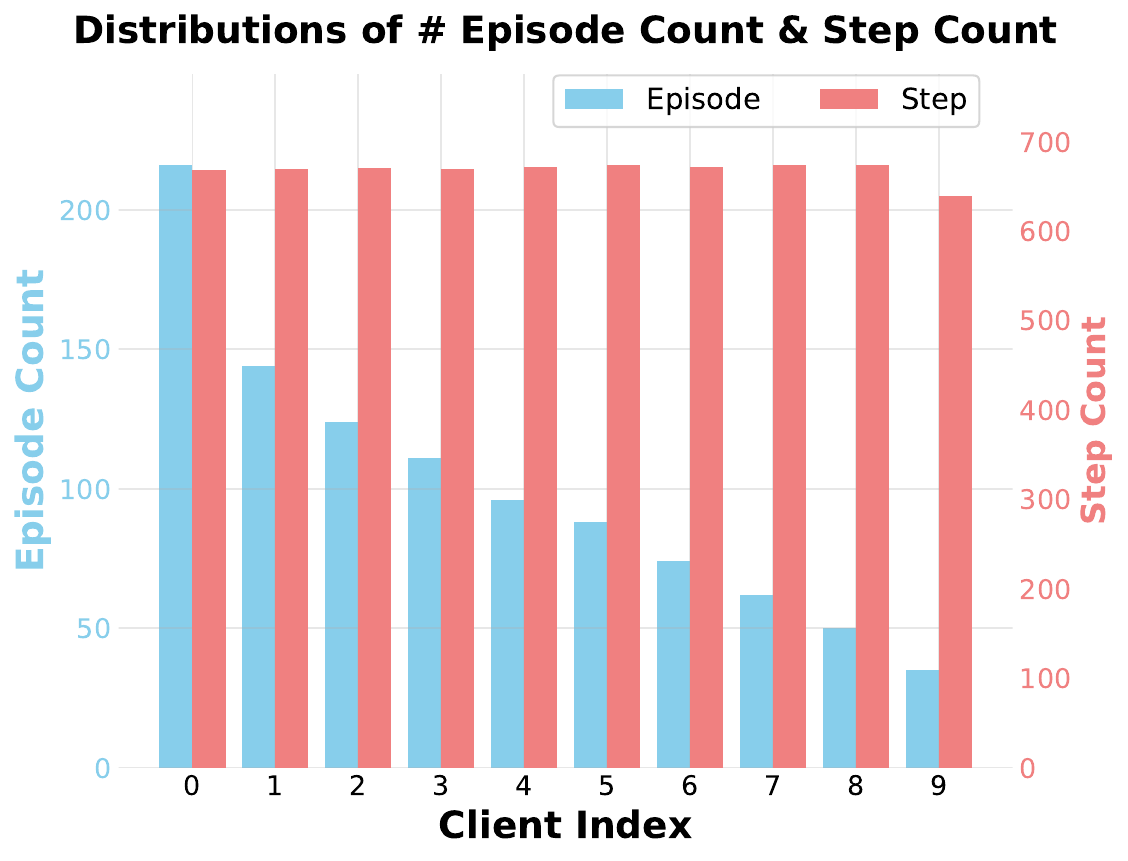}}
		\centerline{(c) Episode Skew}
	\end{minipage}
    \hspace{0.005\linewidth} 
	\begin{minipage}{0.24\linewidth}
		\centerline{\includegraphics[width=\textwidth]{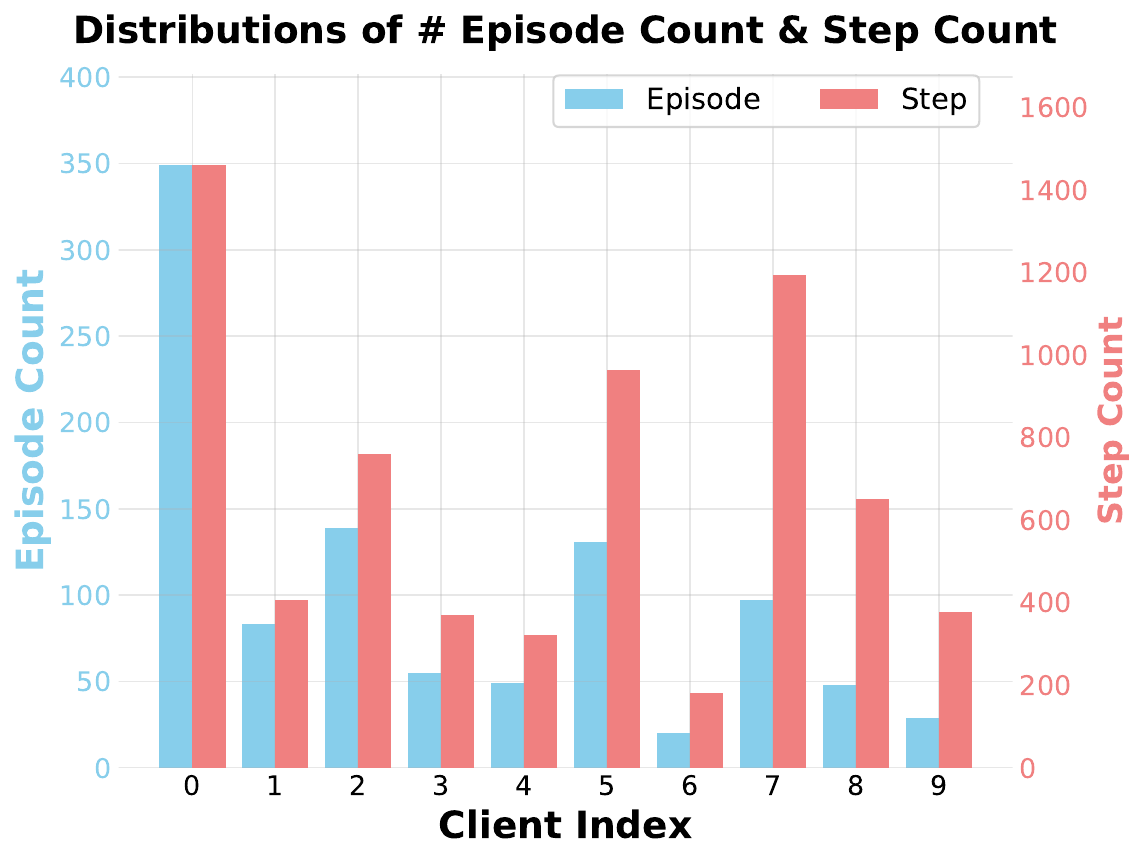}}
		\centerline{(d) Both Skew}
    
	\end{minipage}
\caption{Distributions of episode and step counts within the Step-Episode Dataset. The four subsets highlight distinct differences in average steps per episode across clients. Note that the y-axes are not on the same scale for visualization purposes.
}
\label{fig:dist_step_epi}
\vspace{-1mm}
\end{figure*}

\section{FedMABench}


\subsection{Overview}
FedMABench features with a comprehensive framework and six datasets emphasizing on heterogeneity and diversity.

As shown in Figure \ref{fig:main} (grey), FedMABench adopts the conventional federated learning protocol and provides an easy-to-use, research-friendly framework that includes eight FL baselines. 
Specifically, diverse users with heterogeneous data collaboratively train a global mobile agent on their distributed datasets through four iterative steps: server-to-client model broadcasting, local model training, client-to-server model uploading, and global model aggregation.

In real-world scenarios, mobile users exhibit diverse usage habits and preferences, leading to heterogeneous data distributions which are extremely complex and difficult to quantify. 
To lay the foundation for research on the heterogeneity of distributed data trajectories, we construct two homogeneous datasets and four heterogeneous datasets, addressing diverse aspects of heterogeneity. 
As depicted in Figure \ref{fig:main} (bottom right), our datasets span a wide range of 877 apps across 5 categories, demonstrating another diversity.
A summary of the dataset statistics is presented in Table \ref{tab:summay_datasets}.


\subsection{Data Collection}
\label{sec:data_collection}


\textbf{Data Composition.}
To train the core VLM of mobile agents, each data episode, denoted as \( \mathcal{D} \), comprises multiple steps, each serving as a basic training unit.
A step consists of three components: a task instruction \( \mathcal{T} \), a screenshot, and a corresponding action.  
The data episode is defined as:  
\(
\mathcal{D} = \{\langle \mathcal{T}, a_i, s_i \rangle \mid i \in [1,n] \}  
\), 
where \( \langle \mathcal{T}, a_i, s_i \rangle \) represents the \( i \)-th step, with \( a_i \) and \( s_i \) denoting the action and screenshot respectively.

\textbf{Collection.}
Our datasets are derived from the Android Control and Android in the Wild (AitW) datasets, with two key modifications which are labeling and partitioning. 
Each episode in our datasets is annotated with two app-related attributes: the app name and its corresponding category. 
Given that the original app and category information is not publicly available in \citet{liEffectsDataScale2024}, we are compelled to infer these details based on the actions performed and the instructions provided.
We first employ a dual-strategy method, described in Appendix \ref{app:dataset_details}, to extract the related app name for each episode. 
Following human heuristics, we then categorize the apps into five distinct groups: Shopping, Traveling, Office, Lives, and Entertainment.
We employ GPT-4o to automatically assign each app a corresponding category.
Detailed information regarding the categorization is presented in Table \ref{tab:apps_categorization}.

Subsequently, we partition each constructed dataset into multiple subsets to simulate the federated learning environment, where each subset represents a distinct data distribution.
We specifically control the variables and ensure that subsets are only different in the distribution to provide the fairest possible comparison.

\subsection{Datasets Description}
\label{sec:fedmabench_dataset}
To establish a comprehensive foundation for research, we construct six datasets in FedMABench, emphasizing on different forms of homogeneity or heterogeneity.
This section provides detailed descriptions and visualizations of these datasets, with additional details available in Appendix \ref{app:dataset_details}.

\subsubsection{\textbf{Basic-AC and Basic-AitW Datasets}} ~\\
Initially, we introduce two basic datasets with homogeneous distributions, to validate the general principles and properties of federated learning for training mobile agents.  

\textbf{Description of Basic-AC Dataset.}
The Basic-AC Dataset is constructed based on homogeneous distributions, where we disregard the app attributes of all episodes.
Since all episodes are available in this IID setup, we construct six subordinate training datasets with increasing data sizes: 200, 500, 1,000, 3,000, 5,000, and 7,000 by random sampling. 
Additionally, we create five subsets, each consisting of episodes from a single category, to provide more focused scenarios. 
Basic-AC offers diverse situations with varying data sizes and client participation, enabling the exhaustive evaluation of mobile agents federated trained under IID settings.
For each training set, we sample 10\% of the training size to form the test set.
As shown in Figure \ref{fig:main} (bottom right), Basic-AC covers 877 apps across 5 categories, yielding a diverse distribution. 

\textbf{Description of the Basic-AitW Dataset.}  
To establish a comprehensive experimental foundation with multiple sources, we construct another homogeneous dataset, the Basic-AitW Dataset, derived from AitW. We sample 1,000 episodes from each category to form five subsets, with episodes uniformly assigned to ten clients.
The Basic-AitW dataset offers distinct data characteristics compared to Basic-AC, adding further diversity for benchmarking federated mobile agents.

\begin{figure*}[t]
	\centering
	\begin{minipage}{0.32\linewidth}
		\centerline{\includegraphics[width=\textwidth]{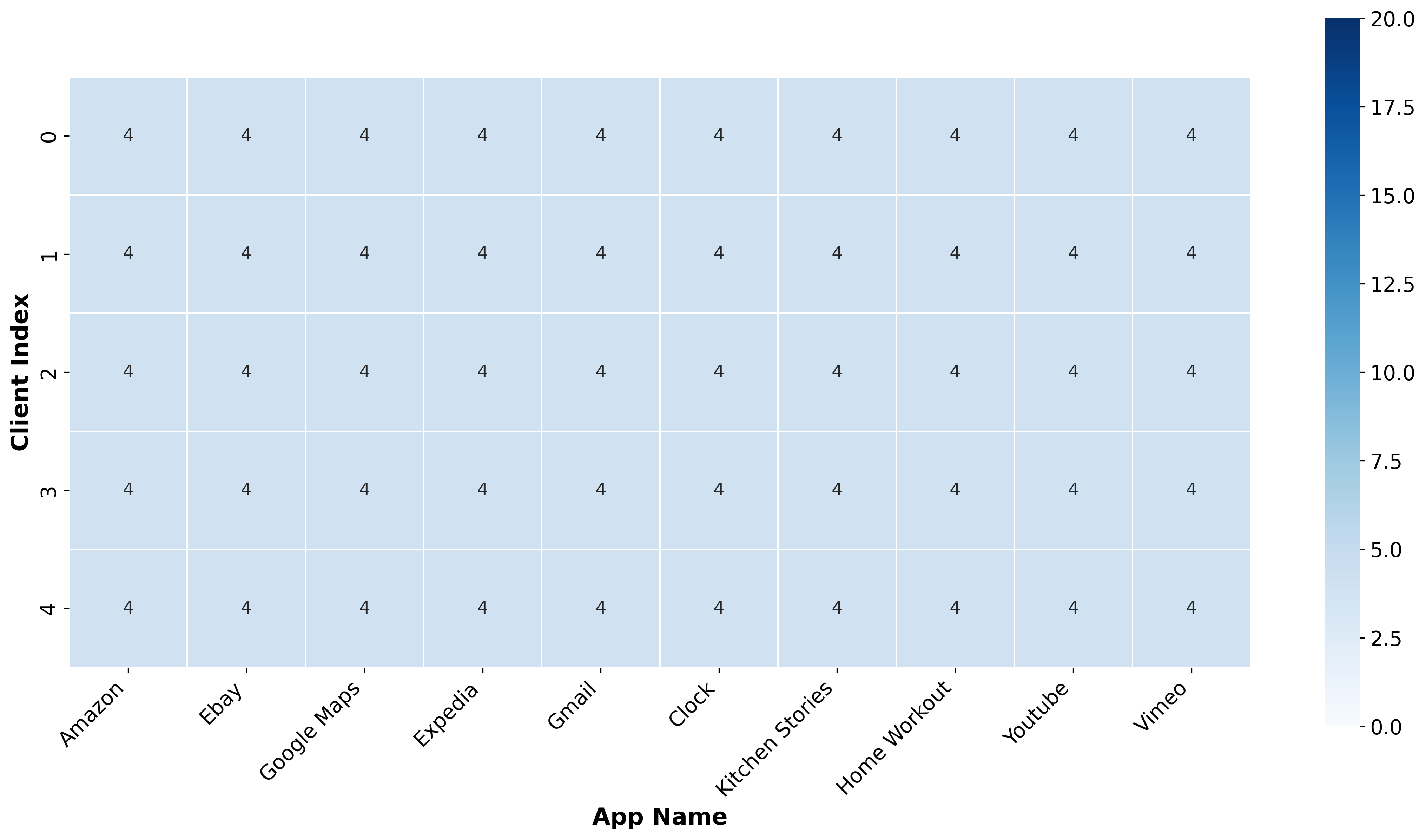}}
		\centerline{(a) Category-App IID}
	\end{minipage}
    \hspace{0.005\linewidth}
    \begin{minipage}{0.32\linewidth}
		\centerline{\includegraphics[width=\textwidth]{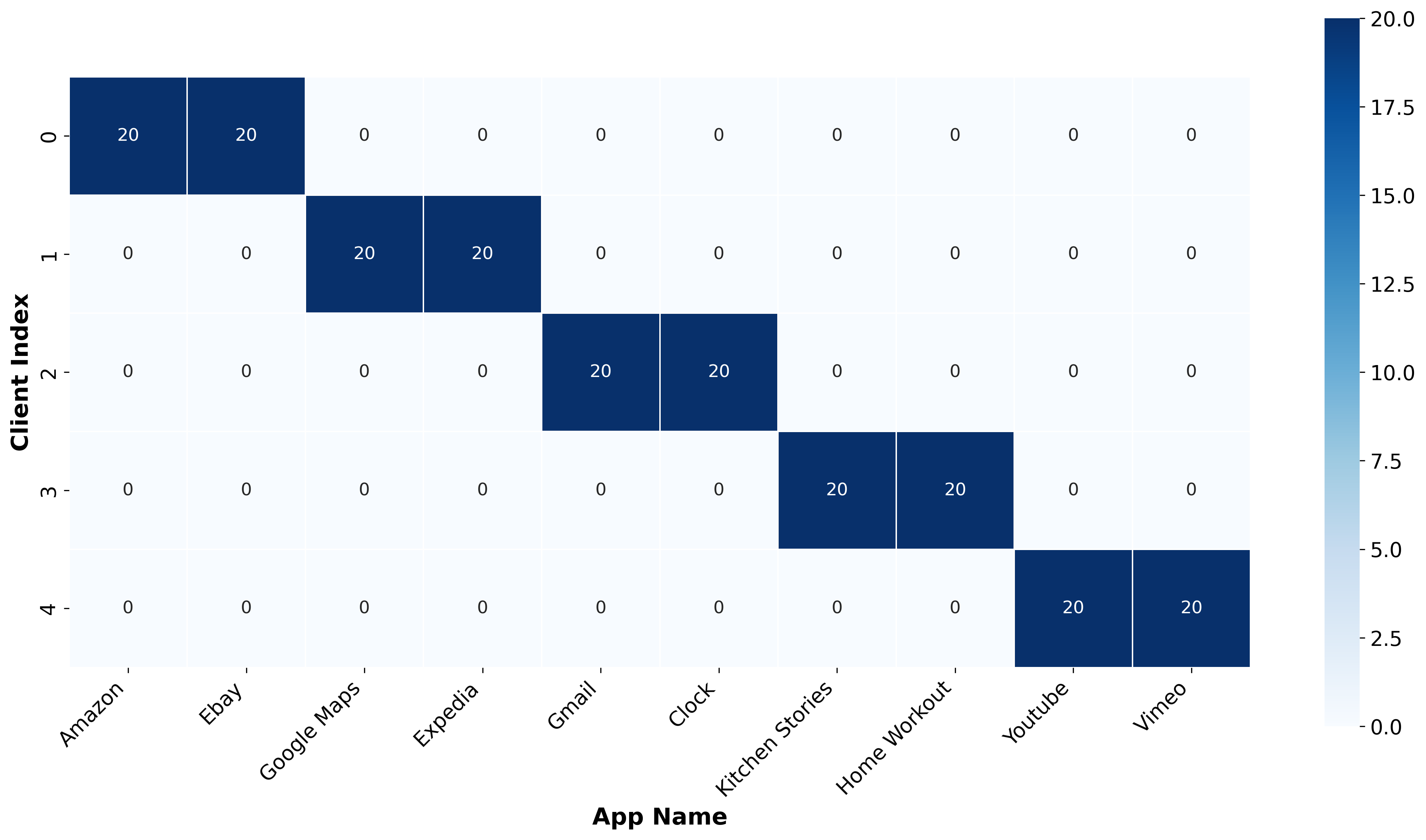}}
		\centerline{(b) Category Skew}
	\end{minipage}
    \hspace{0.005\linewidth}
	\begin{minipage}{0.32\linewidth}
		\centerline{\includegraphics[width=\textwidth]{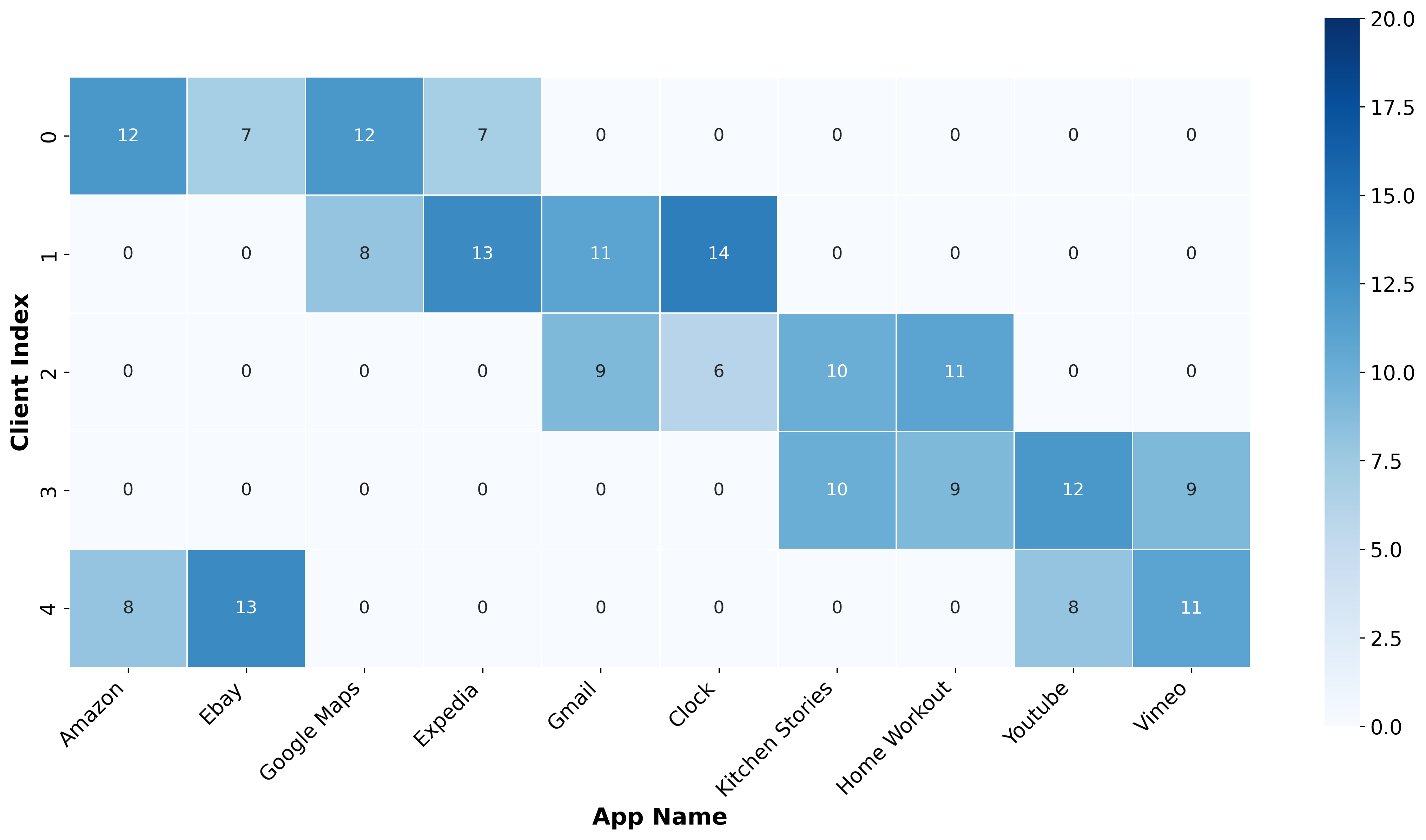}}
		\centerline{(c) Category Half-Skew}
	\end{minipage}
    
	\begin{minipage}{0.32\linewidth}
		\centerline{\includegraphics[width=\textwidth]{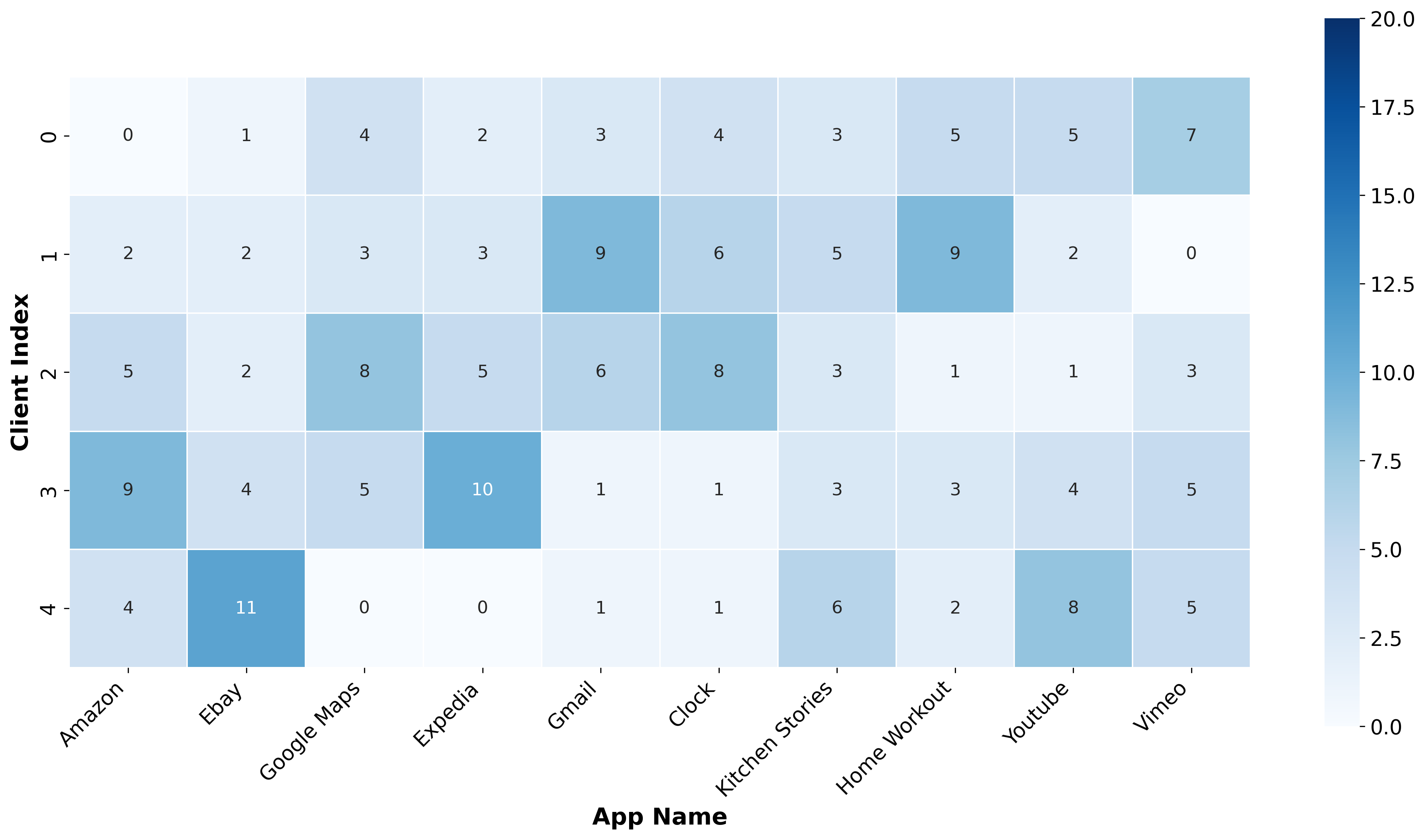}}
		\centerline{(d) Category Non-Uniform}
	\end{minipage}
    \hspace{0.005\linewidth}
    \begin{minipage}{0.32\linewidth}
		\centerline{\includegraphics[width=\textwidth]{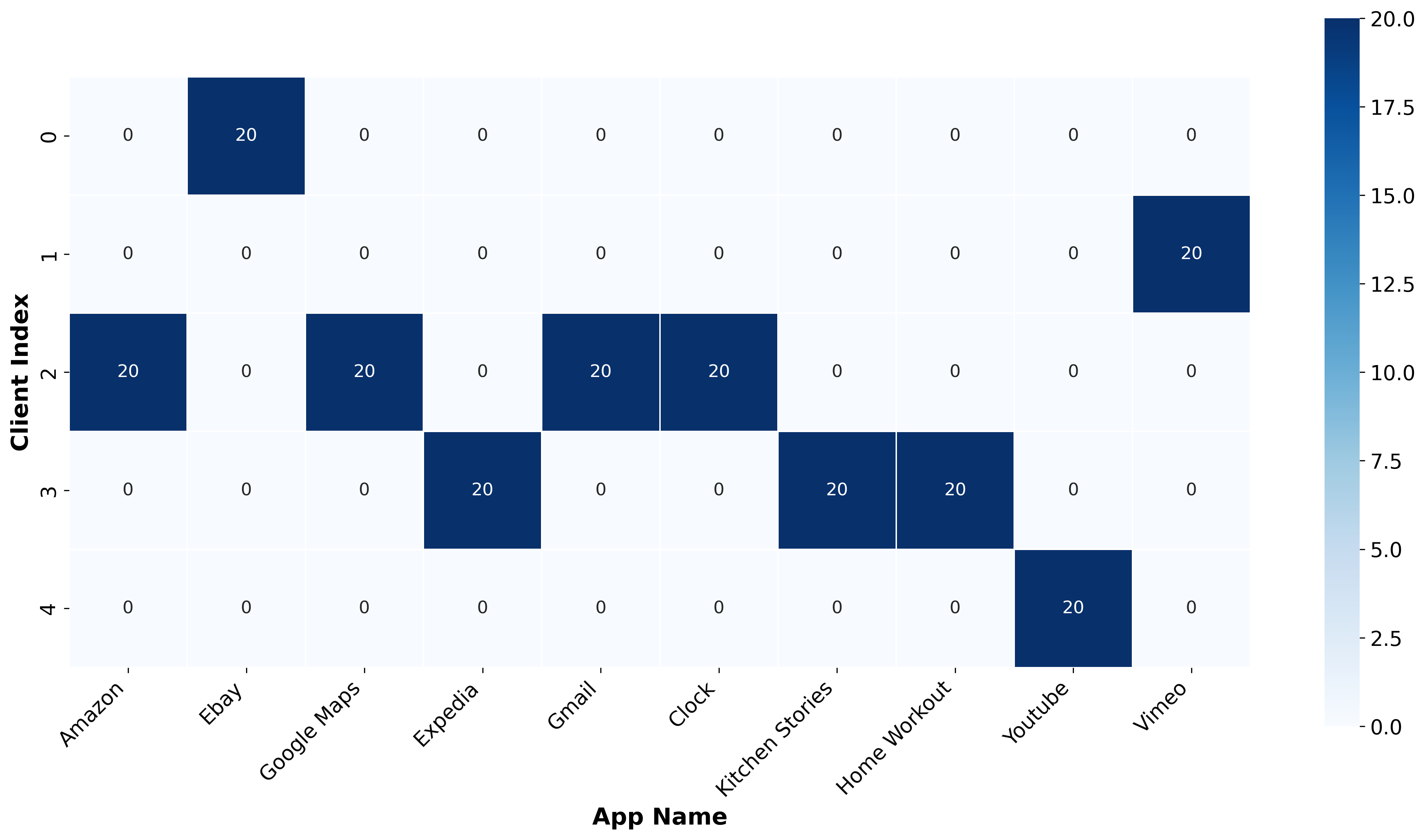}}
		\centerline{(e) App Random}
	\end{minipage}
    \hspace{0.005\linewidth}
	\begin{minipage}{0.32\linewidth}
		\centerline{\includegraphics[width=\textwidth]{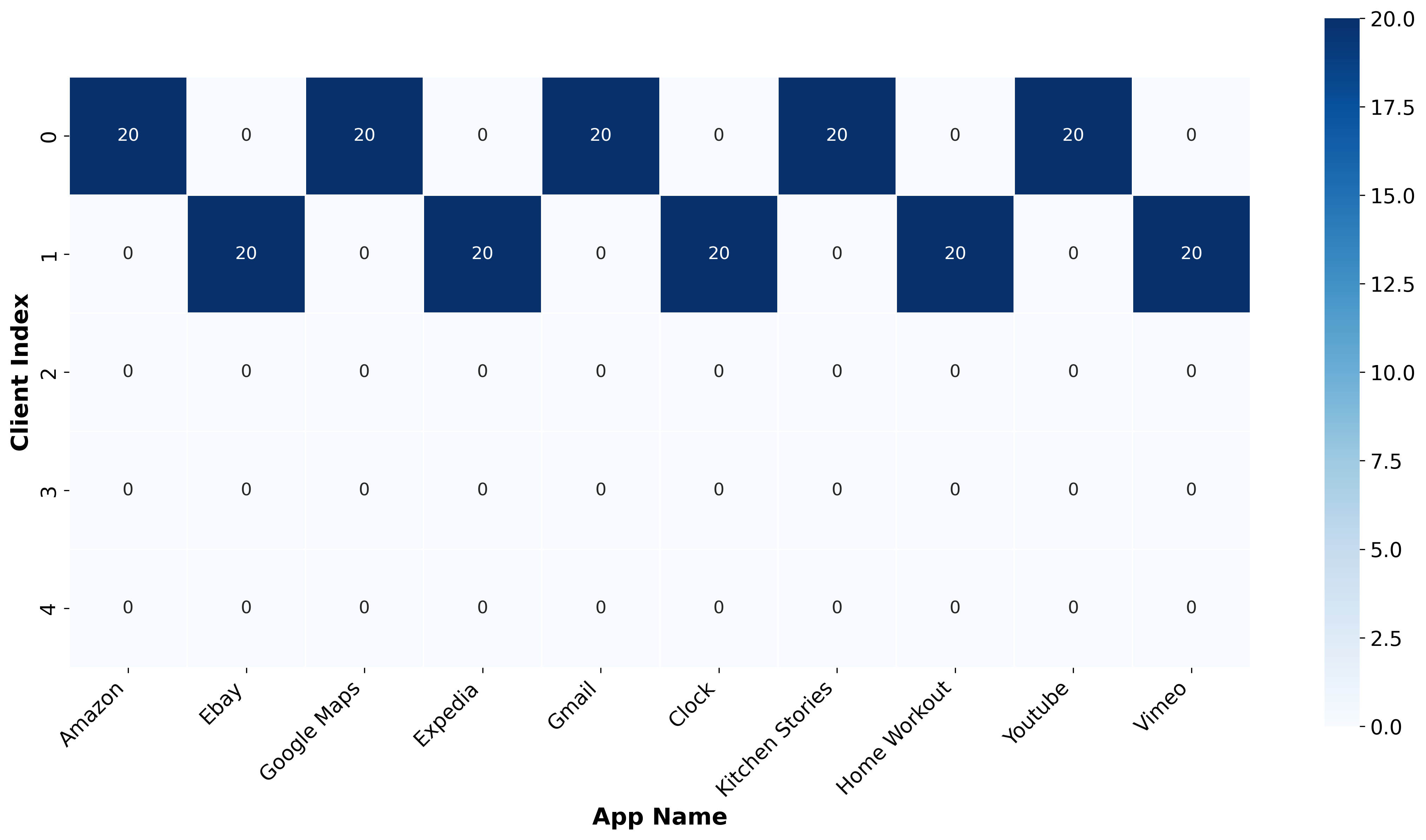}}
		\centerline{(f) App Skew}
	\end{minipage}
\vspace{-1mm}
\caption{
Distributions of the top 10 apps across five clients in the Category-Level Dataset. The top two apps from each of the five categories are selected.  
Our six subsets exhibit diverse patterns in terms of the apps and categories assigned to each client.
}
\label{fig:dist_category}
\vspace{-2mm}
\end{figure*}

\subsubsection{\textbf{Step-Episode Dataset}} ~\\
We construct a series of four datasets, collectively called the Step-Episode Dataset, to promote further research on the heterogeneity of step counts and episode counts across clients.



\textbf{Step-Episode Two-Level Heterogeneity.}
As pointed out in FedMobileAgent \cite{wang2025fedmobileagenttrainingmobileagents}, the distributed user data for training mobile agents exhibits heterogeneity at two levels: step counts and episode counts, due to variances in users' app usage habits.
Specifically, the datasets for training mobile agents are organized by episodes, with each step within an episode serving as the basic training unit. Unlike traditional federated learning tasks, such as image classification or sentiment analysis, the datasets for training federated mobile agents are characterized by two types of quantity measurements: one based on episode counts and the other based on step counts.  
As usage habits vary across different users, these two types of measurements do not necessarily align, leading to a unique form of heterogeneity that cannot be easily evaluated from a single "sample count" perspective, as is typically done in traditional FL tasks. This is why we refer to this heterogeneity as "step-episode two-level".

\textbf{Description \& Visualization.}
To evaluate federated mobile agents under step-episode two-level heterogeneity, we specifically design a series of four subsets.  
To ensure a fair comparison, we base the datasets on the same data pool, which is split into 10 clients following different partition rules.  
To minimize the influence of other types of heterogeneity, such as app usage, we randomly sample from the data pool to form the Step-Episode Dataset. 
The four subsets are as follows:
(1) \textbf{Step-Episode IID}: In this IID subset, all clients have identical step counts and episode counts.
(2) \textbf{Episode Skew}: Clients share similar total step counts, but exhibit skewed episode counts.
(3) \textbf{Step Skew}: All clients have the same episode count, but distinct total step counts.
(4) \textbf{Both Skew}: Both episode and step counts are heterogeneous across clients.

As shown in Figure \ref{fig:dist_step_epi}, the four subsets have diverse step and episode counts which are useful for evaluating mobile agents trained on diverse distributions.

\subsubsection{\textbf{Category-Level Dataset}} ~\\
\label{sec:category}
As users use mobile phones for different purposes, their datasets have heterogeneous app category distributions. Therefore, we construct 6 subsets with controlled variance in distributions to present an empirical study on this heterogeneity.

\textbf{App Category Heterogeneity.}
In real-world user phone usage, the users have various app using habits. As showcased in Figure \ref{fig:main} (grey), some users such as "User 1", use mobile phones mostly for shopping and traveling needs, while others such as "User 2" may often utilize phones for office needs.
Such using habits and needs result in heterogeneous training data for federated mobile agents as the category distributions differ among users.

\textbf{Description \& Visualization.}
To investigate how mobile agents based on classic FL methods perform, we sample 1,000 episodes from the Basic-AC Dataset to form the Category-Level Dataset which consists of 5 categories with 52 apps.
To control and monitor the influence of different apps, we select only those apps with a large number of episodes for research efficiency.  
For a fair comparison, we build six subsets with different partition among five clients and ensure each client has equally 200 episodes.

The sub-datasets are as follows:
(1) \textbf{IID}: In this scenario, each app is evenly allocated across all five clients, meaning each client has the same number of episodes for every app and app category.
(2) \textbf{Category Skew}: In this scenario, the distribution of app categories is highly skewed, as each client possesses only one unique category.
(3) \textbf{Category Half-Skew}: Similar to Category Skewed, in this subset each client has access to two categories, with an even distribution over the two seen categories.
(4) \textbf{Category Non-Uniform}: In this subset, all clients have seen all five categories, but the distribution of categories varies across clients.
(5) \textbf{App Skew}: Each client has five categories of apps, but within each category, a particular app is only seen by one client. In other words, the category distribution is IID across clients, but the specific apps within each category are completely different.
(6) \textbf{App Random}: Each app is only seen by one client, with apps randomly assigned to clients.

We choose the top 2 apps for each category and plot their distribution heatmaps in Figure \ref{fig:dist_category}. The figures clearly highlight the notable distinctions between subsets.

\begin{figure*}[t]
	\centering
	\begin{minipage}{0.24\linewidth}
		\centerline{\includegraphics[width=\textwidth]{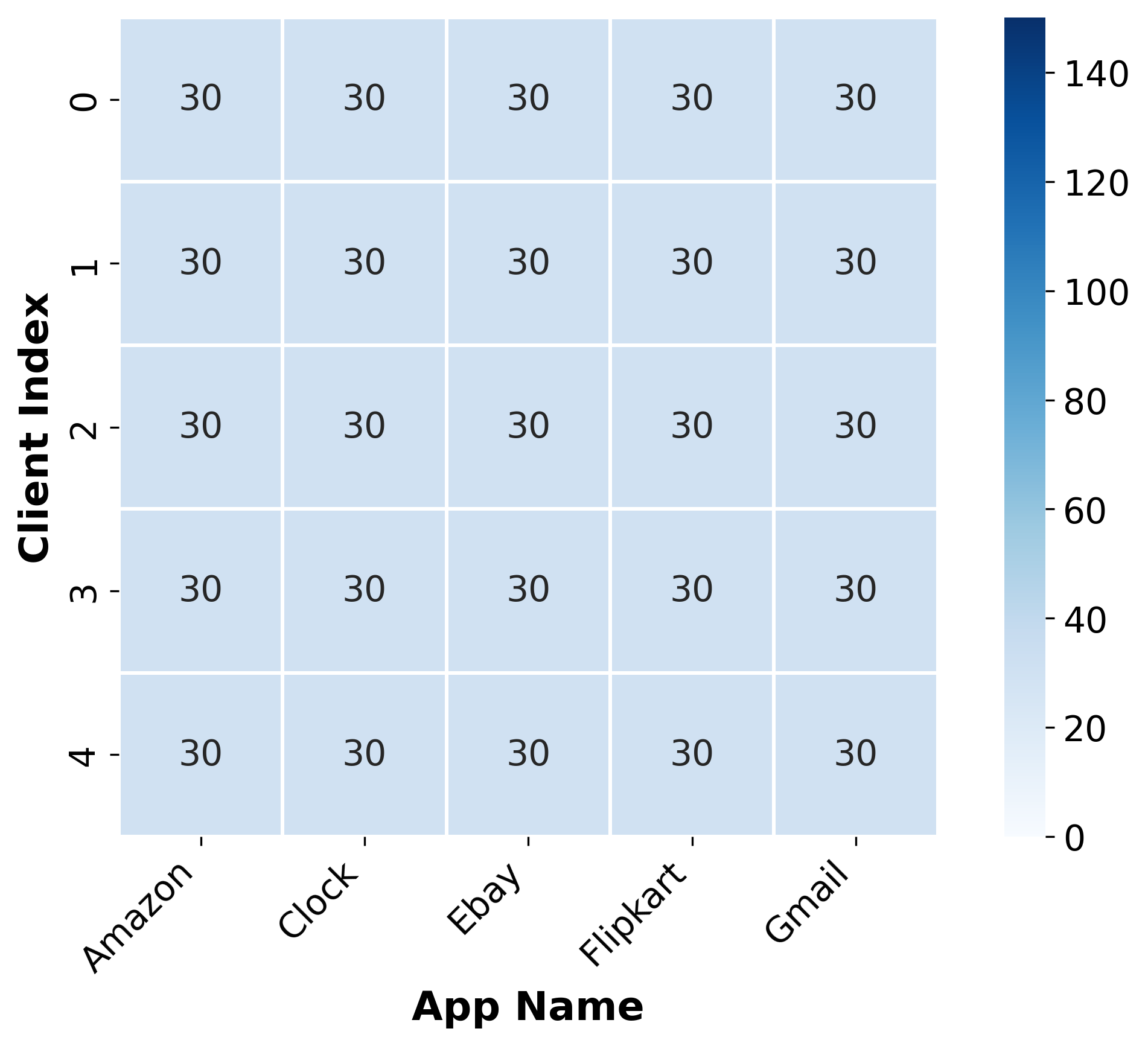}}
		\centerline{(a) App IID}
	\end{minipage}
    \hspace{0.004\linewidth}
    \begin{minipage}{0.24\linewidth}
		\centerline{\includegraphics[width=\textwidth]{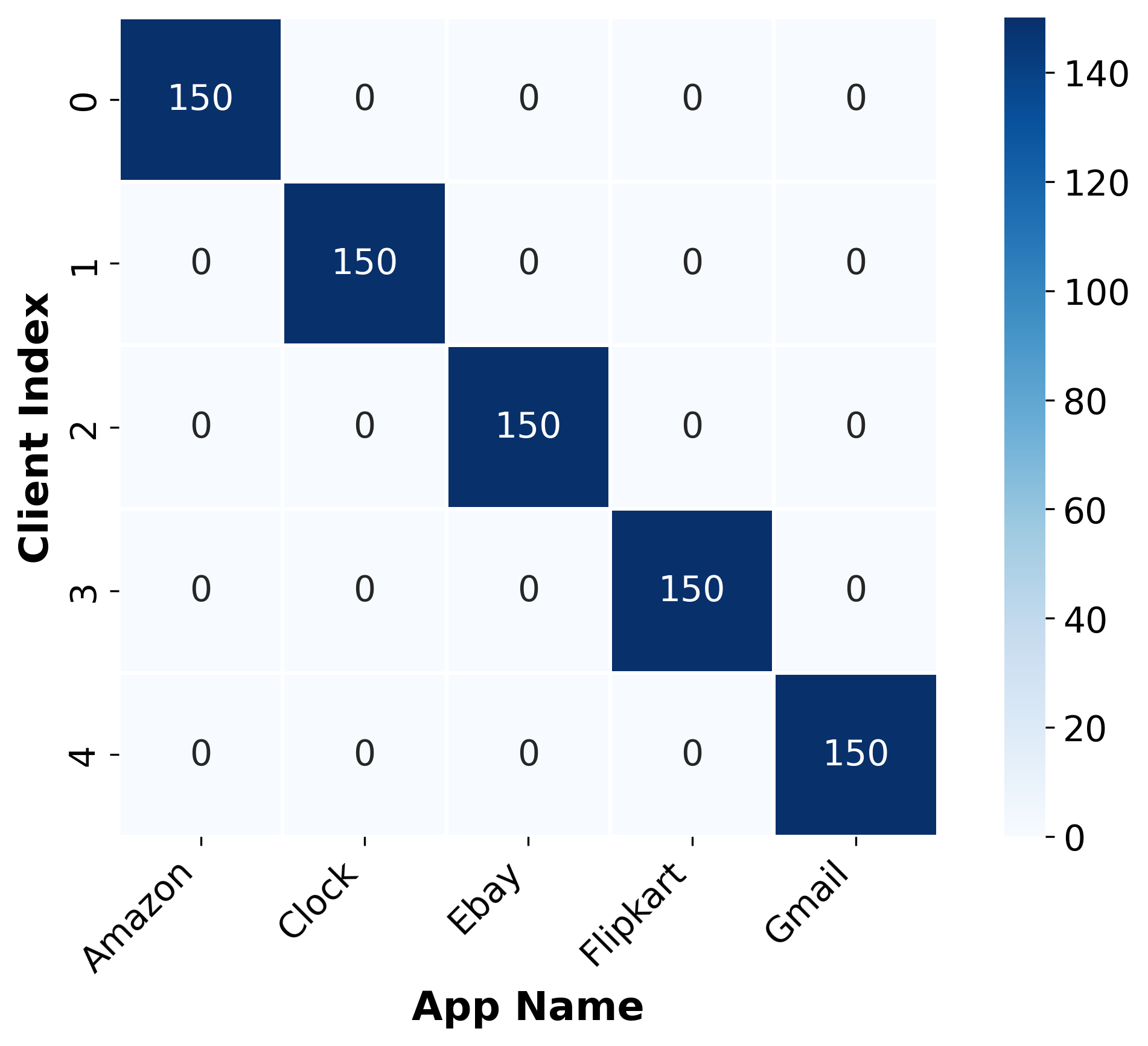}}
		\centerline{(b) App Skew}
	\end{minipage}
    \hspace{0.004\linewidth}
	\begin{minipage}{0.24\linewidth}
		\centerline{\includegraphics[width=\textwidth]{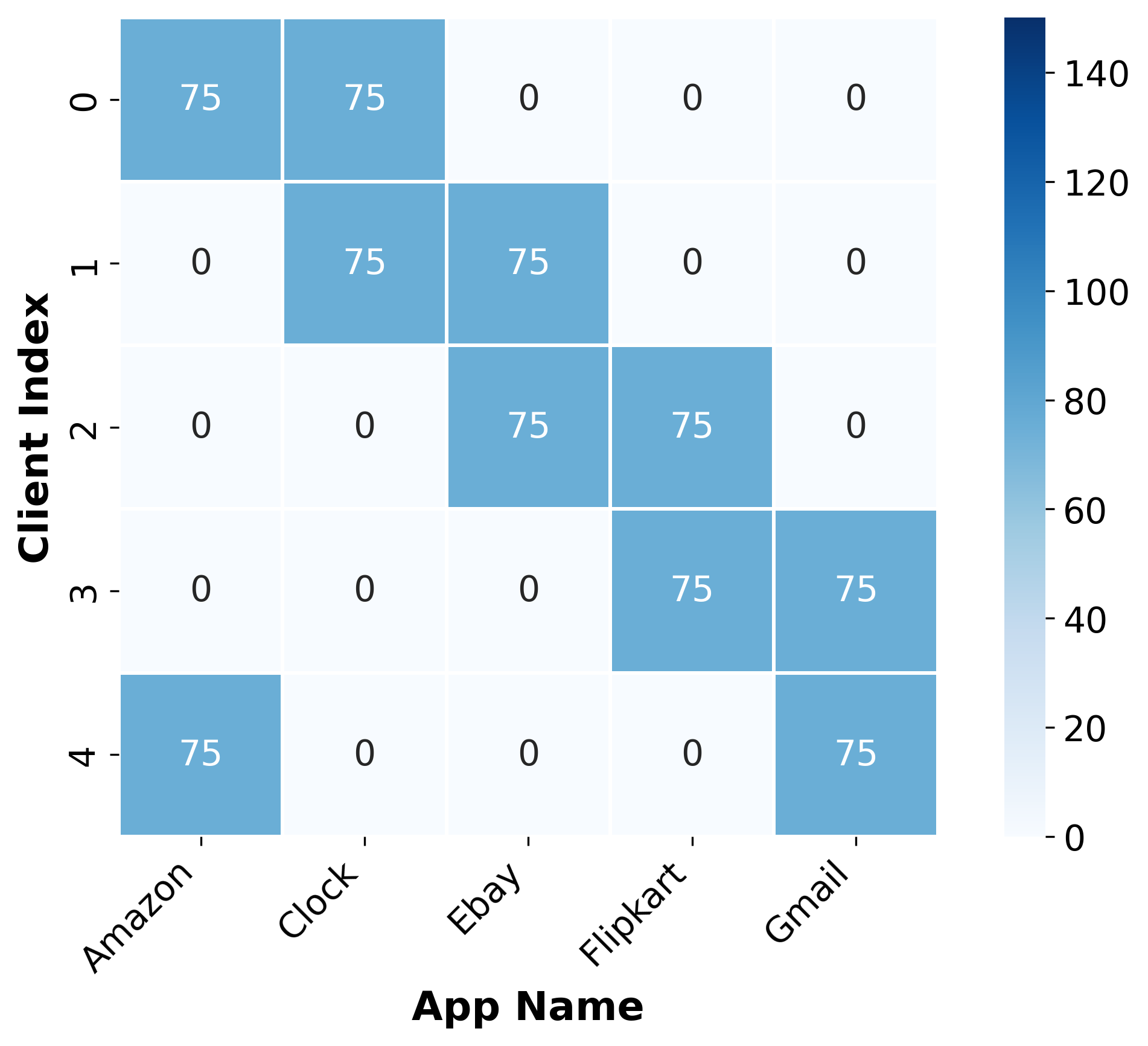}}
		\centerline{(c) App Half-Skew}
	\end{minipage}
    \hspace{0.004\linewidth}
	\begin{minipage}{0.24\linewidth}
		\centerline{\includegraphics[width=\textwidth]{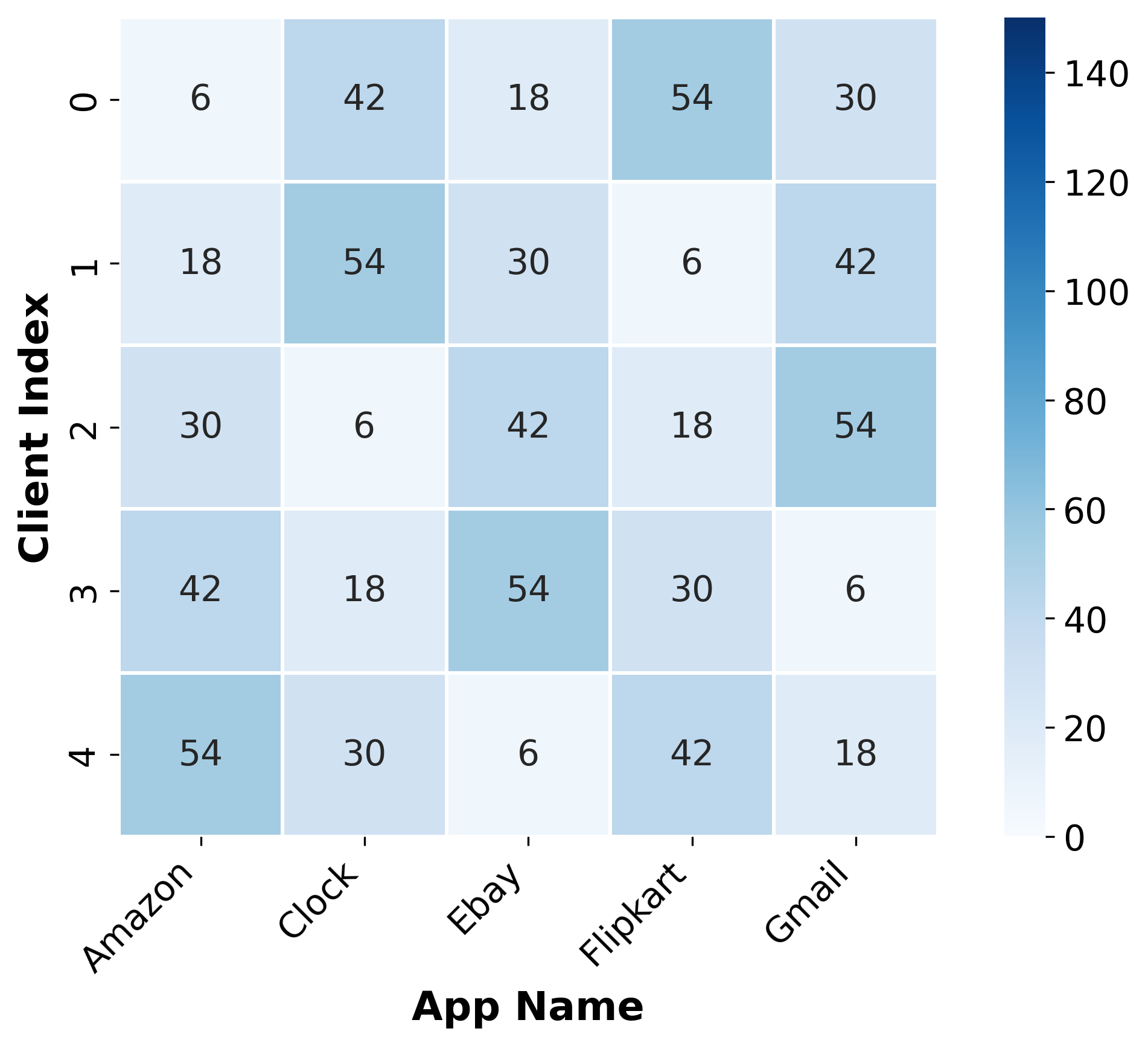}}
		\centerline{(d) Non-Uniform}
	\end{minipage}
\vspace{-1mm}
\caption{
Distributions of the five apps across the App-Level Dataset. Our subsets reveal distinct differences in the heterogeneity of app usage. Note that the numbers represent episode counts, and the episodes are identical for all subsets.
}
\label{fig:dist_app}
\vspace{-2mm}
\end{figure*}

\subsubsection{\textbf{App-Level and ScaleApp Datasets}} ~\\
To evaluate on the app-level heterogeneity instead of categories, we build a concise dataset called App-Level Dataset targeted at 5 apps and another dataset with scaled app and client numbers.

\textbf{App Name Heterogeneity.}
In real life, mobile phone users exhibit distinct preferences for specific apps, even among those that serve similar functions.
Therefore, this form of heterogeneity cannot be measured from the perspective of app categories, but rather by app names.
As showcased in Figure \ref{fig:main}, "User K" prefers Amazon over eBay for purchasing products and Epic over Steam for gaming, leading to heterogeneity in the specific apps used.

\textbf{Description of App-Level Dataset.}
We construct a series of datasets aimed at capturing this diversity in apps. 
To make the distinction more apparent and straightforward for research comparison, we select five apps with the highest usage frequencies: Amazon, Clock, eBay, Flipkart, and Gmail. Given the limitations in available data samples for each individual app, we sample 150 episodes for each app. 
Subsequently, we create four representative subsets following similar insights to those described in Section \ref{sec:category}:
(1) \textbf{App-IID:} All clients share the same number of episodes for each app.
(2) \textbf{App Skew:} Each client has data collected from only one specific app.
(3) \textbf{App Half-Skew:} Each client has access to two apps with an equal distribution of episodes.
(4) \textbf{App Non-Uniform:} All clients have seen all five apps but with varying distributions of data.
To facilitate comprehensive research, we provide a test dataset with an equal number of episodes for each of the five apps.

\textbf{Description of ScaleApp Dataset.}
Furthermore, we extend the App-Level Dataset to create ScaleApp, a series of scaled datasets designed to incorporate a larger number of clients and apps, thereby better emulating real-world scenarios with a more diverse user base. 
In this extended setup, we select 30 apps, yielding a total of 2,500 episodes distributed across 30 clients, to explore the impact of varying user behaviors and app distributions.

We construct three subsets, where clients with the same index have the same number of episodes, but the number varies across different indexes, which differs from the configuration in the App-Level Dataset:
(1) \textbf{ScaleApp IID}: All clients are exposed to all apps, with uniform distribution across clients.
(2) \textbf{ScaleApp Skew}: Each client only possesses data belong to one specific app.
(3) \textbf{ScaleApp Random}: The episodes are randomly allocated to the 30 clients.

\textbf{Visualization.}
The distribution heatmaps for the App-Level and ScaleApp datasets are shown in Figures \ref{fig:dist_app} and \ref{fig:hetmap_ScaleApp}.  
From the figures, we observe diverse distribution patterns between different subsets, with minimal influence from unrelated variables due to tight control, making them suitable for effective research comparisons.

\vspace{-2.5mm}
\subsection{Framework Description}
\label{sec:framework_des}
FedMABench integrates eight typical federated learning algorithms and supports more than ten base models. Our supported models and implemented methods are summarized in Table \ref{tab:models_algs}.
In addition, we establish an end-to-end pipeline that offers two training settings: high-level and low-level training, each can be evaluated using two metrics: step-level accuracy and episode-level accuracy.  
Unlike high-level training, low-level training provides the agent with the subgoal as an extra input for the current step.
We build our training and evaluation framework based on the well-known repository \texttt{ms-swift} \cite{zhao2024swiftascalablelightweightinfrastructure}. It is important to note that incorporating federated learning support is non-trivial, as we decompose the training pipeline and successfully integrate federated training in a concise manner, which facilitates the easy reproduction of other algorithms.  

\begin{table}[t]
\centering
\small
\caption{Supported base models and integrated FL algorithms.}
\vspace{-1mm}
\begin{tabular}{ll}
\toprule
\multicolumn{2}{l}{\textbf{Supported Base Models}} \\
Qwen Families \cite{Qwen2VL}: & Qwen2-VL-2B/7B-Instruct, Qwen-VL-Chat \\ 
Intern Families \cite{chen2024internvl}: & InternVL2-1B/2B/4B/8B \\
DeepSeek Families \cite{lu2024deepseekvl}: & DeepSeekVL2, DeepSeekVL2-tiny/small \\
OpenAI Families \cite{gpt4}: & GPT-4o, GPT-4o-mini, GPT-4-Vision  \\ 
\midrule
\multicolumn{2}{l}{\textbf{Integrated FL Algorithms}} \\
\multicolumn{2}{l}{FedAvg\cite{fedavg}, FedProx\cite{fedprox}, SCAFFOLD\cite{scaffold}, FedAvgM\cite{fedavgm},} \\
\multicolumn{2}{l}{FedAdam, FedYogi, FedAdagrad \cite{fedopt}, FedMobileAgent\cite{wang2025fedmobileagenttrainingmobileagents} } \\
\bottomrule
\end{tabular}

\label{tab:models_algs}
\vspace{-2mm}
\end{table}

\begin{table*}[t]
\centering
\caption{Experiments on the Basic-AC and Basic-AitW Datasets. FedAvg consistently surpasses Local Learning, validating the effectiveness of training mobile agents on distributed user data. 
Local 0 denotes the \(0\)-th client as a representative of all clients.}
\vspace{-2mm}
\begin{tabular}{l|ccc|cccccc}
\toprule
\multirow{3}{*}{\textbf{Algorithm}} & \multicolumn{3}{c|}{\textbf{Basic-AC}} & \multicolumn{6}{c}{\textbf{Basic-AitW}}  \\
 ~ & \textbf{High-Level} & \multicolumn{2}{c|}{\textbf{Low-Level}} & \multirow{2}{*}{General} & \multirow{2}{*}{Install} & \multirow{2}{*}{G-Apps} & \multirow{2}{*}{Single} & \multirow{2}{*}{WebShopping} & \multirow{2}{*}{\textbf{Avg.}} \\
~ & Step Acc & Step Acc & Episode Acc & \multicolumn{6}{c}{~} \\
\midrule
Zero-Shot & 27.24 & 52.13 & 6 & 15.90 & 5.20 & 15.08 & 28.38 & 11.41 & 15.19 \\ 
Central & 55.59 & 80.47 & 27 & 35.04 & 54.50 & 46.65 & 55.46 & 39.82 & 46.29 \\ 
Local 0 & 37.64 & 70.87 & 20 & 35.21 & 52.47 & 36.03 & 45.41 & 32.04 & 40.23 \\ 
FedAvg & 50.87 & 78.90 & 33 & 36.56 & 51.84 & 38.27 & 54.59 & 33.59 & 42.97 \\ 
\bottomrule
\end{tabular}
\label{tab:basic}
\vspace{-2mm}
\end{table*}

\begin{table}[t]
\centering
\caption{Experiments with multiple baselines on the Step-Episode Dataset. In this setting, FedMobileAgent achieves best performance on average and outperforms GPT-4o, one of the SOTA VLMs. Best results are denoted in \textbf{bold}.
}
\vspace{-2mm}
\begin{tabular}{l|cccc|>{\columncolor{gray!20}}c}
\toprule
\textbf{Algorithm} & IID & Episode & Step & Both & \cellcolor{white}\textbf{Avg.}  \\ 
\midrule
Qwen2-7B-Instruct & \multicolumn{4}{c|}{\cellcolor{gray!10}27.24} & 27.24 \\
GPT-4o & \multicolumn{4}{c|}{\cellcolor{gray!10}42.52} & 42.52 \\
Central & \multicolumn{4}{c|}{\cellcolor{gray!10}55.59} & 55.59 \\ 
\midrule
Local 0 & \cellcolor{red!10}37.64 & \cellcolor{blue!10}33.39 & \cellcolor{blue!10}29.13 & \cellcolor{blue!10}\textbf{46.77} & 36.73  \\ 
FedAvg & \cellcolor{red!10}\textbf{43.78} & \cellcolor{blue!10}40.63 & \cellcolor{blue!10}40.63 & \cellcolor{blue!10}40.81 & 41.46  \\ 
FedProx & \cellcolor{red!10}42.36 & \cellcolor{blue!10}41.10 & \cellcolor{blue!10}40.16 & \cellcolor{blue!10}40.16 & 40.95  \\ 
FedAvgM & \cellcolor{red!10}42.00 & \cellcolor{blue!10}41.57 & \cellcolor{blue!10}41.10 & \cellcolor{blue!10}40.47 & 41.29  \\ 
FedYogi & \cellcolor{red!10}42.05 & \cellcolor{blue!10}41.10 & \textbf{\cellcolor{blue!10}41.26} & \cellcolor{blue!10}42.05 & 41.62  \\ 
FedAdagrad & \cellcolor{red!10}43.31 & \cellcolor{blue!10}41.42 & \cellcolor{blue!10}41.10 & \cellcolor{blue!10}41.26 & 41.77  \\ 
SCAFFOLD & \cellcolor{red!10}41.73 & \cellcolor{blue!10}41.42 & \textbf{\cellcolor{blue!10}41.26} & \cellcolor{blue!10}39.84 & 41.06  \\ 
FedMobileAgent & \cellcolor{red!10}42.68 & \cellcolor{blue!10}\textbf{41.89} & \cellcolor{blue!10}\textbf{41.26} & \cellcolor{blue!10}46.53 & \textbf{43.09} \\  

\bottomrule

\end{tabular}
\label{tab:hetero_step_episode}
\vspace{-3mm}
\end{table}



\begin{table*}[t]
\centering
\setlength\tabcolsep{3pt}
\caption{Experiments on the Category-Level Dataset. Skewed category distribution results in slightly lower accuracy. FL algorithms exhibit diverse behaviors. Entertain. is short for Entertainment. Colors represent \colorbox{red!10}{homogeneity} and \colorbox{blue!10}{heterogeneity}. }
\vspace{-1.5mm}
\begin{tabular}{l|cccccc|l|cccccc}
\toprule
\textbf{Algorithm} & Shopping & Traveling & Office & Lives & Entertain. & \textbf{Avg.} & \textbf{Algorithm} & Shopping & Traveling & Office & Lives & Entertain. & \textbf{Avg.} \\
\midrule
Zero-Shot & \cellcolor{gray!10}26.61 & \cellcolor{gray!10}25.33 & \cellcolor{gray!10}27.05 & \cellcolor{gray!10}24.41 & \cellcolor{gray!10}23.81 & \cellcolor{gray!20}25.46 & Central & \cellcolor{gray!10}57.26 & \cellcolor{gray!10}58.67 & \cellcolor{gray!10}51.64 & \cellcolor{gray!10}55.12 & \cellcolor{gray!10}60.95 & \cellcolor{gray!20}56.90 \\ 
\midrule

\textbf{Homo.} & \multicolumn{6}{c|}{\cellcolor{red!10}\textbf{Category IID}} & \textbf{Hetero.} & \multicolumn{6}{c}{\cellcolor{blue!10}\textbf{Category Skew}}\\
Local 0 & \cellcolor{red!10}48.39 & \cellcolor{red!10}45.78 & \cellcolor{red!10}36.89 & \cellcolor{red!10}32.28 & \cellcolor{red!10}45.71 & \cellcolor{red!20}42.25 & Local 0 & \cellcolor{blue!10}50.81 & \cellcolor{blue!10}47.56 & \cellcolor{blue!10}46.72 & \cellcolor{blue!10}38.58 & \cellcolor{blue!10}48.57 & \cellcolor{blue!20}46.51 \\ 
FedAvg & \cellcolor{red!10}\textbf{55.65} & \cellcolor{red!10}52.00 & \cellcolor{red!10}52.46 & \cellcolor{red!10}37.80 & \cellcolor{red!10}\textbf{51.43} & \cellcolor{red!20}50.07 & FedAvg & \cellcolor{blue!10}52.42 & \cellcolor{blue!10}52.00 & \cellcolor{blue!10}\textbf{48.36} & \cellcolor{blue!10}41.73 & \cellcolor{blue!10}\textbf{52.38} & \cellcolor{blue!20}49.64 \\ 
FedProx & \cellcolor{red!10}53.23 & \cellcolor{red!10}52.44 & \cellcolor{red!10}51.64 & \cellcolor{red!10}\textbf{38.58} & \cellcolor{red!10}\textbf{51.43} & \cellcolor{red!20}49.79 & FedProx & \cellcolor{blue!10}51.61 & \cellcolor{blue!10}52.44 & \cellcolor{blue!10}47.54 & \cellcolor{blue!10}41.73 & \cellcolor{blue!10}49.52 & \cellcolor{blue!20}49.08 \\ 
FedAvgM & \cellcolor{red!10}\textbf{54.84} & \cellcolor{red!10}52.89 & \cellcolor{red!10}50.00 & \cellcolor{red!10}\textbf{38.58} & \cellcolor{red!10}49.52 & \cellcolor{red!20}49.64 & FedAvgM & \cellcolor{blue!10}\textbf{54.84} & \cellcolor{blue!10}52.89 & \cellcolor{blue!10}\textbf{48.36} & \cellcolor{blue!10}\textbf{42.52} & \cellcolor{blue!10}\textbf{52.38} & \cellcolor{blue!20}\textbf{50.50} \\ 
FedYogi & \cellcolor{red!10}54.84 & \cellcolor{red!10}\textbf{53.78} & \cellcolor{red!10}\textbf{52.46} & \cellcolor{red!10}\textbf{38.58} & \cellcolor{red!10}50.48 & \cellcolor{red!20}\textbf{50.50} & FedYogi & \cellcolor{blue!10}54.03 & \cellcolor{blue!10}\textbf{53.78} & \cellcolor{blue!10}\textbf{48.36} & \cellcolor{blue!10}41.73 & \cellcolor{blue!10}51.43 & \cellcolor{blue!20}50.36 \\
\midrule

\textbf{Hetero.} & \multicolumn{6}{c|}{\cellcolor{blue!10}\textbf{Category Half-Skew}} & \textbf{Hetero.} & \multicolumn{6}{c}{\cellcolor{blue!10}\textbf{Category Non-Uniform}}\\
Local 0 & \cellcolor{blue!10}41.13 & \cellcolor{blue!10}\textbf{56.00} & \cellcolor{blue!10}36.89 & \cellcolor{blue!10}\textbf{40.16} & \cellcolor{blue!10}37.14 & \cellcolor{blue!20}44.38 & Local 0 & \cellcolor{blue!10}38.71 & \cellcolor{blue!10}33.78 & \cellcolor{blue!10}34.43 & \cellcolor{blue!10}34.65 & \cellcolor{blue!10}33.33 & \cellcolor{blue!20}34.85 \\ 
FedAvg & \cellcolor{blue!10}46.77 & \cellcolor{blue!10}47.11 & \cellcolor{blue!10}39.34 & \cellcolor{blue!10}36.22 & \cellcolor{blue!10}\textbf{47.62} & \cellcolor{blue!20}43.81 & FedAvg & \cellcolor{blue!10}\textbf{50.00} & \cellcolor{blue!10}48.89 & \cellcolor{blue!10}47.54 & \cellcolor{blue!10}40.94 & \cellcolor{blue!10}46.67 & \cellcolor{blue!20}47.08 \\ 
FedProx & \cellcolor{blue!10}\textbf{47.58} & \cellcolor{blue!10}49.33 & \cellcolor{blue!10}\textbf{42.62} & \cellcolor{blue!10}39.37 & \cellcolor{blue!10}46.67 & \cellcolor{blue!20}\textbf{45.66} & FedProx & \cellcolor{blue!10}47.94 & \cellcolor{blue!10}\textbf{52.42} & \cellcolor{blue!10}\textbf{50.22} & \cellcolor{blue!10}\textbf{45.90} & \cellcolor{blue!10}42.52 & \cellcolor{blue!20}46.67 \\ 
FedAvgM & \cellcolor{blue!10}45.16 & \cellcolor{blue!10}48.89 & \cellcolor{blue!10}\textbf{42.62} & \cellcolor{blue!10}38.58 & \cellcolor{blue!10}45.71 & \cellcolor{blue!20}44.81 & FedAvgM & \cellcolor{blue!10}48.39 & \cellcolor{blue!10}51.56 & \cellcolor{blue!10}46.72 & \cellcolor{blue!10}43.31 & \cellcolor{blue!10}\textbf{48.57} & \cellcolor{blue!20}\textbf{48.22} \\ 
FedYogi & \cellcolor{blue!10}43.55 & \cellcolor{blue!10}46.22 & \cellcolor{blue!10}36.07 & \cellcolor{blue!10}34.65 & \cellcolor{blue!10}40.00 & \cellcolor{blue!20}40.97 & FedYogi & \cellcolor{blue!10}46.77 & \cellcolor{blue!10}52.00 & \cellcolor{blue!10}47.54 & \cellcolor{blue!10}43.31 & \cellcolor{blue!10}\textbf{48.57} & \cellcolor{blue!20}\textbf{48.22} \\

\bottomrule
\end{tabular}
\label{tab:hetero_category}
\vspace{-1mm}
\end{table*}

\begin{table*}[t]
\centering
\setlength\tabcolsep{4.7pt}
\caption{Experiments on the App-Level Dataset. We provide evaluation results on all five apps. FL algorithms in skewed app distributions perform significantly lower accuracy compared to IID situations. 
}
\vspace{-1.5mm}
\begin{tabular}{l|cccccc|l|cccccc}
\toprule
\textbf{Algorithm} & Amazon & Clock & Ebay & Flipkart & Gmail & \textbf{Avg.} & \textbf{Algorithm} & Amazon & Clock & Ebay & Flipkart & Gmail & \textbf{Avg.} \\ 
\midrule
Zero-Shot & \cellcolor{gray!10}29.75 & \cellcolor{gray!10}32.38 & \cellcolor{gray!10}28.33 & \cellcolor{gray!10}30.00 & \cellcolor{gray!10}28.12 & \cellcolor{gray!20}29.62 & Central & \cellcolor{gray!10}54.55 & \cellcolor{gray!10}64.76 & \cellcolor{gray!10}58.33 & \cellcolor{gray!10}61.00 & \cellcolor{gray!10}51.56 & \cellcolor{gray!20}57.67 \\
\midrule
\textbf{Homo.} & \multicolumn{6}{c|}{\cellcolor{red!10}\textbf{App IID}} & \textbf{Hetero.} & \multicolumn{6}{c}{\cellcolor{blue!10}\textbf{App Skew}}\\
Local 0 & \cellcolor{red!10}44.63 & \cellcolor{red!10}49.52 & \cellcolor{red!10}41.67 & \cellcolor{red!10}50.00 & \cellcolor{red!10}33.59 & \cellcolor{red!20} \cellcolor{red!20}43.38 & Local 0 & \cellcolor{blue!10}\textbf{56.20} & \cellcolor{blue!10}36.19 & \cellcolor{blue!10}42.50 & \cellcolor{blue!10}44.00 & \cellcolor{blue!10}21.09 & \cellcolor{blue!20}39.72 \\
Local 1 & \cellcolor{red!10}46.28 & \cellcolor{red!10}\textbf{57.14} & \cellcolor{red!10}52.50 & \cellcolor{red!10}54.00 & \cellcolor{red!10}39.06 & \cellcolor{red!20} \cellcolor{red!20}49.30 & Local 1 & \cellcolor{blue!10}33.06 & \cellcolor{blue!10}\textbf{60.00} & \cellcolor{blue!10}38.33 & \cellcolor{blue!10}31.00 & \cellcolor{blue!10}28.91 & \cellcolor{blue!20}37.80 \\
Local 2 & \cellcolor{red!10}54.55 & \cellcolor{red!10}53.33 & \cellcolor{red!10}51.67 & \cellcolor{red!10}51.00 & \cellcolor{red!10}38.28 & \cellcolor{red!20} \cellcolor{red!20}49.48 & Local 2 & \cellcolor{blue!10}40.50 & \cellcolor{blue!10}17.14 & \cellcolor{blue!10}45.00 & \cellcolor{blue!10}37.00 & \cellcolor{blue!10}20.31 & \cellcolor{blue!20}32.06 \\
FedAvg & \cellcolor{red!10}57.02 & \cellcolor{red!10}53.33 & \cellcolor{red!10}52.50 & \cellcolor{red!10}55.00 & \cellcolor{red!10}46.88 & \cellcolor{red!20} \cellcolor{red!20}52.79 & FedAvg & \cellcolor{blue!10}48.76 & \cellcolor{blue!10}53.33 & \cellcolor{blue!10}\textbf{48.33} & \cellcolor{blue!10}52.00 & \cellcolor{blue!10}42.97 & \cellcolor{blue!20}48.78 \\
FedProx & \cellcolor{red!10}55.37 & \cellcolor{red!10}53.33 & \cellcolor{red!10}\textbf{55.00} & \cellcolor{red!10}54.00 & \cellcolor{red!10}44.53 & \cellcolor{red!20} \cellcolor{red!20}52.26 & FedProx & \cellcolor{blue!10}48.76 & \cellcolor{blue!10}53.33 & \cellcolor{blue!10}\textbf{48.33} & \cellcolor{blue!10}\textbf{54.00} & \cellcolor{blue!10}39.84 & \cellcolor{blue!20}48.43 \\
FedAvgM & \cellcolor{red!10}\textbf{58.68} & \cellcolor{red!10}52.38 & \cellcolor{red!10}54.17 & \cellcolor{red!10}54.00 & \cellcolor{red!10}46.88 & \cellcolor{red!20} \cellcolor{red!20}53.14 & FedAvgM & \cellcolor{blue!10}49.59 & \cellcolor{blue!10}53.33 & \cellcolor{blue!10}\textbf{48.33} & \cellcolor{blue!10}52.00 & \cellcolor{blue!10}39.84 & \cellcolor{blue!20}48.26 \\
FedYogi & \cellcolor{red!10}57.02 & \cellcolor{red!10}54.29 & \cellcolor{red!10}54.17 & \cellcolor{red!10}\textbf{58.00} & \cellcolor{red!10}\textbf{48.44} & \cellcolor{red!20} \cellcolor{red!20}\textbf{54.18} & FedYogi & \cellcolor{blue!10}48.76 & \cellcolor{blue!10}54.29 & \cellcolor{blue!10}47.50 & \cellcolor{blue!10}\textbf{54.00} & \cellcolor{blue!10}\textbf{43.75} & \cellcolor{blue!20}\textbf{49.30} \\
\midrule
\textbf{Hetero.} & \multicolumn{6}{c|}{\cellcolor{blue!10}\textbf{App Half-Skew}} & \textbf{Hetero.} & \multicolumn{6}{c}{\cellcolor{blue!10}\textbf{App Non-Uniform}}  \\
Local 0 & \cellcolor{blue!10}52.89 & \cellcolor{blue!10}\textbf{57.14} & \cellcolor{blue!10}45.00 & \cellcolor{blue!10}40.00 & \cellcolor{blue!10}36.72 & \cellcolor{blue!20}46.17 & Local 0 & \cellcolor{blue!10}39.67 & \cellcolor{blue!10}\textbf{58.10} & \cellcolor{blue!10}38.33 & \cellcolor{blue!10}48.00 & \cellcolor{blue!10}\textbf{46.09} & \cellcolor{blue!20}45.64 \\ 
Local 1 & \cellcolor{blue!10}\textbf{57.02} & \cellcolor{blue!10}53.33 & \cellcolor{blue!10}\textbf{50.00} & \cellcolor{blue!10}47.00 & \cellcolor{blue!10}28.91 & \cellcolor{blue!20}46.86 & Local 1 & \cellcolor{blue!10}52.89 & \cellcolor{blue!10}56.19 & \cellcolor{blue!10}38.33 & \cellcolor{blue!10}47.00 & \cellcolor{blue!10}39.84 & \cellcolor{blue!20}46.52 \\ 
Local 2 & \cellcolor{blue!10}50.41 & \cellcolor{blue!10}40.95 & \cellcolor{blue!10}41.67 & \cellcolor{blue!10}\textbf{58.00} & \cellcolor{blue!10}28.91 & \cellcolor{blue!20}41.64 & Local 2 & \cellcolor{blue!10}47.11 & \cellcolor{blue!10}49.52 & \cellcolor{blue!10}45.00 & \cellcolor{blue!10}\textbf{55.00} & \cellcolor{blue!10}40.62 & \cellcolor{blue!20}47.04 \\ 
FedAvg & \cellcolor{blue!10}54.55 & \cellcolor{blue!10}53.33 & \cellcolor{blue!10}45.83 & \cellcolor{blue!10}55.00 & \cellcolor{blue!10}38.28 & \cellcolor{blue!20}48.95 & FedAvg & \cellcolor{blue!10}56.20 & \cellcolor{blue!10}55.24 & \cellcolor{blue!10}45.83 & \cellcolor{blue!10}51.00 & \cellcolor{blue!10}42.19 & \cellcolor{blue!20}49.83 \\ 
FedProx & \cellcolor{blue!10}56.20 & \cellcolor{blue!10}55.24 & \cellcolor{blue!10}43.33 & \cellcolor{blue!10}55.00 & \cellcolor{blue!10}38.28 & \cellcolor{blue!20}49.13 & FedProx & \cellcolor{blue!10}\textbf{57.02} & \cellcolor{blue!10}55.24 & \cellcolor{blue!10}45.83 & \cellcolor{blue!10}50.00 & \cellcolor{blue!10}38.28 & \cellcolor{blue!20}48.95 \\ 
FedAvgM & \cellcolor{blue!10}54.55 & \cellcolor{blue!10}53.33 & \cellcolor{blue!10}45.00 & \cellcolor{blue!10}54.00 & \cellcolor{blue!10}\textbf{42.19} & \cellcolor{blue!20}\textbf{49.48} & FedAvgM & \cellcolor{blue!10}55.37 & \cellcolor{blue!10}54.29 & \cellcolor{blue!10}45.83 & \cellcolor{blue!10}50.00 & \cellcolor{blue!10}41.41 & \cellcolor{blue!20}49.13 \\ 
FedYogi & \cellcolor{blue!10}54.55 & \cellcolor{blue!10}51.43 & \cellcolor{blue!10}44.17 & \cellcolor{blue!10}55.00 & \cellcolor{blue!10}41.41 & \cellcolor{blue!20}48.95 & FedYogi & \cellcolor{blue!10}55.37 & \cellcolor{blue!10}55.24 & \cellcolor{blue!10}\textbf{46.67} & \cellcolor{blue!10}52.00 & \cellcolor{blue!10}42.19 & \cellcolor{blue!20}\textbf{50.00} \\ 

\bottomrule
\end{tabular}
\label{tab:app_level}
\vspace{-2mm}
\end{table*}

\vspace{-1mm}
\section{Experiments}
We conduct extensive experiments on all six datasets using the FL algorithms implemented in FedMABench.
\vspace{-2mm}
\subsection{Basic Setups}
\label{sec:exp_basic_setup}
Note: more experimental details are attached in Appendix \ref{app:ExperimentalDetails}.

\textbf{Model Architecture.}
We employ Qwen2-VL-7B-Instruct \cite{Qwen2VL} as the base model for most of our experiments. We use Low-Rank Adaptation (LoRA) \cite{hu2021lora} for efficient fine-tuning as the resources are limited on mobile phones.
If not otherwise mentioned, Zero-Shot denotes the result of Qwen2-VL-7B-Instruct.

\textbf{Training Configuration.}
We train every model for 10 rounds and sample 10\% the total dataset at each round.
In most settings, we randomly sample 3 clients to participate each round to simulate real-world scenarios where users are occasionally offline \cite{dropout}.

\textbf{Evaluation Protocol.}
We propose a two-tier evaluation framework to assess agent performance:
Step Accuracy measures precision at the action level by checking if the predicted response matches the ground truth based on TF-IDF similarity.
Episode Accuracy evaluates task execution success, requiring all steps in an episode to be correct.

\vspace{-1mm}
\subsection{Experiments on Basic-AC \& Basic-AitW}
\textbf{Setups.}
In this section, the experiments are based on the two homogeneous datasets of FedMABench to examine the general properties of federated learning in training mobile agents. 
From all available subsets we choose those with 1,000 episodes as representatives. We evaluated four methodologies on behalf of all baselines, using step-level accuracy as the primary evaluation metric.
For Basic-AC, we perform both high-level training and low-level training. Since the episode accuracies of high-level training are close, we omit them for brevity.
For Basic-AitW, we experiment on each subset separately and provide the average results as well. The evaluation sets are consistent with those used in FedMobileAgent \cite{wang2025fedmobileagenttrainingmobileagents}.

\textbf{Results.}
From Table \ref{tab:basic}, we draw the following conclusions:
(1) Federated learning effectively leverages distributed user data, as evidenced by the noticeable improvement of FedAvg over local training on both the Basic-AC and Basic-AitW Datasets. However, the performance of FedAvg still falls short of centralized training, which aligns with expectations.
(2) Federated learning yields varying levels of improvement across different subsets of Basic-AitW, highlighting the impact of different data types and laying the foundation for exploring heterogeneity in the following sections.

\vspace{-1mm}
\subsection{Experiments on Step-Episode Dataset}
\label{sec:exp_step_episode}
\textbf{Setups.}
We compare seven baselines and two base models on all four subsets: Step-Episode IID, Episode Skew, Step Skew and Both Skew (short for IID, Episode, Step and Both in Table \ref{tab:hetero_step_episode} respectively). 
The evaluation dataset is consistent to provide straightforward comparison, which is why the results are identical for Centralized learning and base models across subsets.
Note that we intentionally evaluate FedMobileAgent with the parameter \(\lambda\) set to 7 (around the average steps per episode), which is designed to balance the two-level heterogeneity in both step and episode counts.

\textbf{Results.}
As shown in Table \ref{tab:hetero_step_episode}, the results indicate that:
(1) The presence of two-level heterogeneity in step and episode counts is evident, as there is a clear performance drop when the federated trained mobile agents shift from IID scenarios to other non-IID scenarios.
(2) Different federated learning algorithms exhibit distinct behaviors in response to this heterogeneity. Overall, FedMobileAgent\cite{wang2025fedmobileagenttrainingmobileagents}, which leverages a weighted aggregation of each client’s total steps and episodes, demonstrates the best performance under these heterogeneous conditions. This approach effectively captures the disparities in data contributions across clients, thereby mitigating the performance drop caused by the two-level sample count heterogeneity.
(3) It is surprising at firts sight, that Local 0 performs exceptionally well on the Both Skew subset. However, Figure \ref{fig:dist_step_epi} (d) shows that the \(0\)-th client holds a large portion of the total data, which explains its superior performance.

\vspace{-1.5mm}
\subsection{Experiments on Category-Level Dataset}
\label{sec:exp_category}
\textbf{Setups.}
We construct 6 subsets to examine how federated mobile agents behave with heterogeneous app category distributions.  
Due to page limits, we present 4 subsets in Table \ref{tab:hetero_category}, with the remaining provided in the Appendix (Table \ref{tab:hetero_category_supplement1} and \ref{tab:hetero_category_supplement2}).
The red color and blue color represent homogeneous and heterogeneous datasets respectively.
We evaluate performance across all five category and report the average accuracy across all test samples. 

\textbf{Results.}
In our constructed hierarchy, heterogeneity escalates from mild to severe as we progress from Category IID \(\xrightarrow{}\) Non-Uniform \(\xrightarrow{}\) Half-Skew \(\xrightarrow{}\) Skew. However, the general accuracy results in Table \ref{tab:hetero_category} rank as Category IID \(>\) Skew \(>\) Non-Uniform \(>\) Half-Skew, which does not precisely align with the expected heterogeneity levels.  
These results suggest that:  
(1) App category heterogeneity exists and degrades federated learning performance, as nearly all algorithms show a performance drop when transitioning from homogeneous to heterogeneous scenarios.  
(2) Despite explicit shifts in category distributions, the results on the Category Skew subset remain statistically comparable to those on the Category IID subset. This suggests that category differences lead to domain-invariant representations (i.e., features common across categories, such as temporal usage patterns) which counteract the harmful effects of heterogeneity.
In summary, app category differences are not the fundamental cause of heterogeneity.

\vspace{-1.5mm}
\subsection{Experiments on App-Level Dataset}
\label{sec:exp_app_level}
\textbf{Setups.}
The App-Level Dataset encompasses 5 apps: Amazon, Clock, Ebay, Flipkart and Gmail. We evaluate all 5 apps and report their average performance across four subsets. The color scheme follows the same convention as in Section \ref{sec:exp_category}.  
Additionally, we include more results from training on the \(1\)-st and \(2\)-nd clients to offer more comparative insights and useful findings.

\textbf{Results.}
As shown in Table \ref{tab:app_level}, we conclude the following:
(1) The presence of app heterogeneity is evident, as there is a clear performance drop when the model shifts to heterogeneous situations.
(2) We further observe a positive correlation between the severity of app name heterogeneity and performance degradation, confirming that this form of heterogeneity not only exists but critically impacts model effectiveness in real-world deployment contexts.
(3) In comparison with the results from the Category-Level Dataset, we find that differences in specific app names contribute more significantly to heterogeneity than app categories.
(4) Overall, FedYogi \cite{fedopt} outperforms other representative FL algorithms.
(5) Interestingly, we observe that the \(1\)-st client in the App Half-Skew subset, which only has access to episodes from Clock and Ebay, outperforms all FL baselines on Amazon. We hypothesize that there may be underlying relationships between these apps that warrant further exploration.


\vspace{-2mm}
\section{Conclusions}
In this paper, we present FedMABench, the first research-friendly and comprehensive benchmark for federated learning of mobile agents, accompanied by 6 novel datasets encompassing over 30 meticulously designed subsets that capture representative patterns of real-world heterogeneity. 
Our extensive experiments reveal insightful discoveries, such as differences in specific app names contribute more significantly to heterogeneity than app categories. 
Overall, FedMABench bridges the critical gap between theoretical FL research and practical mobile agent applications, laying a solid foundation for future work.





\bibliographystyle{ACM-Reference-Format}
\bibliography{agent-3}


\appendix
\begin{figure*}[t]
	\centering
	\begin{minipage}{0.32\linewidth}
		\centerline{\includegraphics[width=\textwidth]{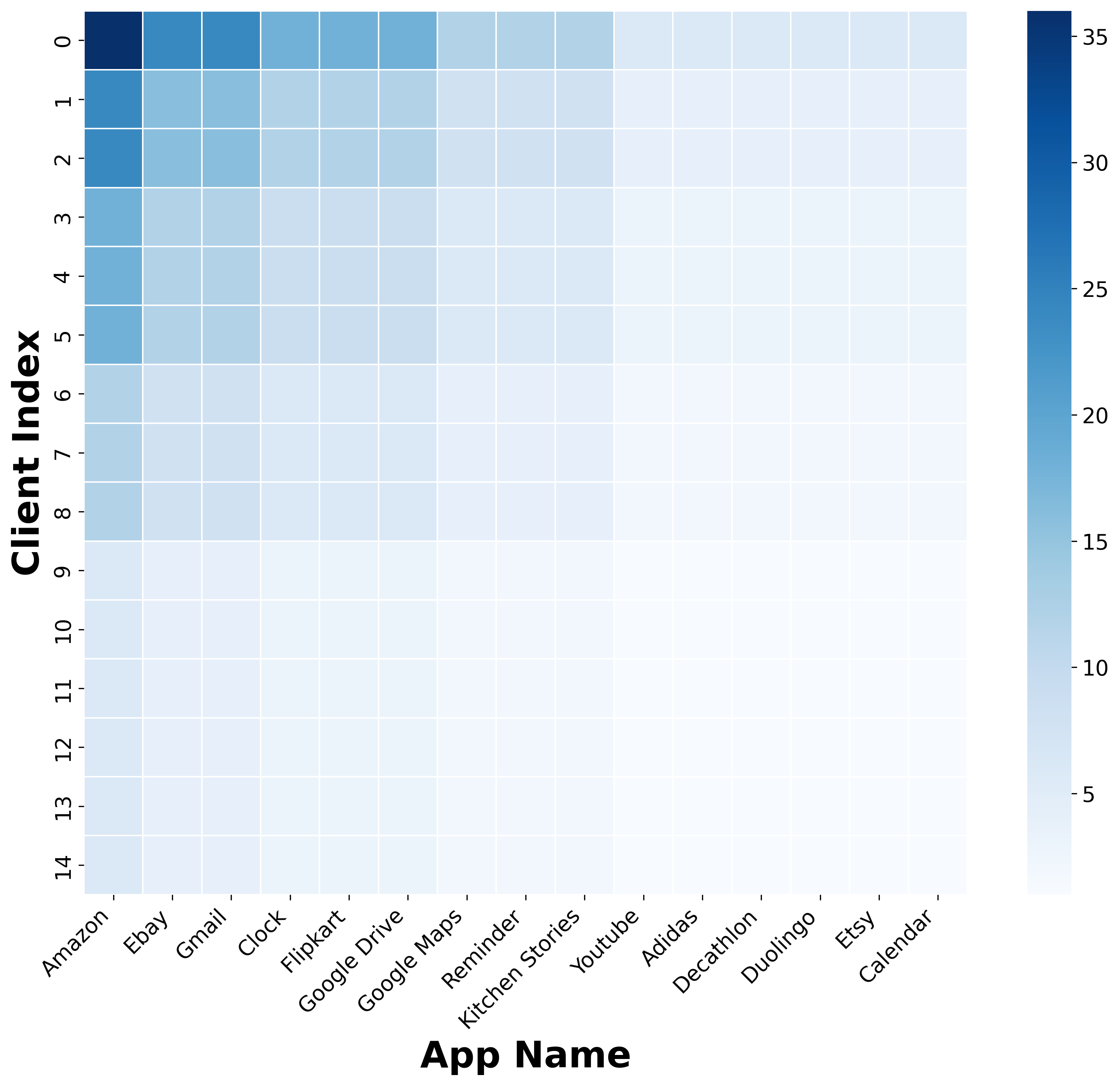}}
		\centerline{(a) ScaleApp IID}
	\end{minipage}
    \hspace{0.005\linewidth}
    \begin{minipage}{0.32\linewidth}

		\centerline{\includegraphics[width=\textwidth]{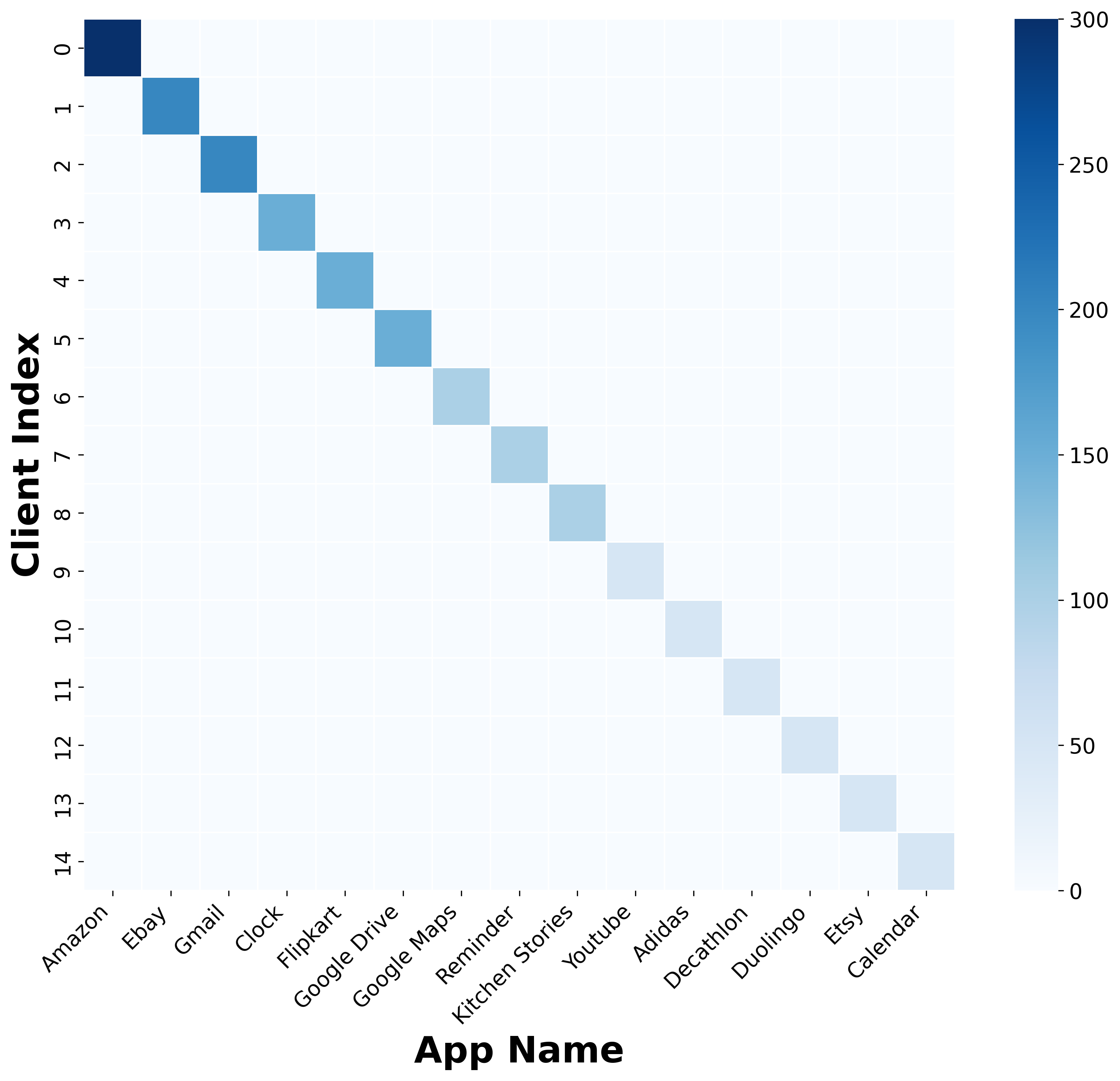}}
		\centerline{(b) ScaleApp Skew}
	\end{minipage}
    \hspace{0.005\linewidth}
	\begin{minipage}{0.32\linewidth}
		\centerline{\includegraphics[width=\textwidth]{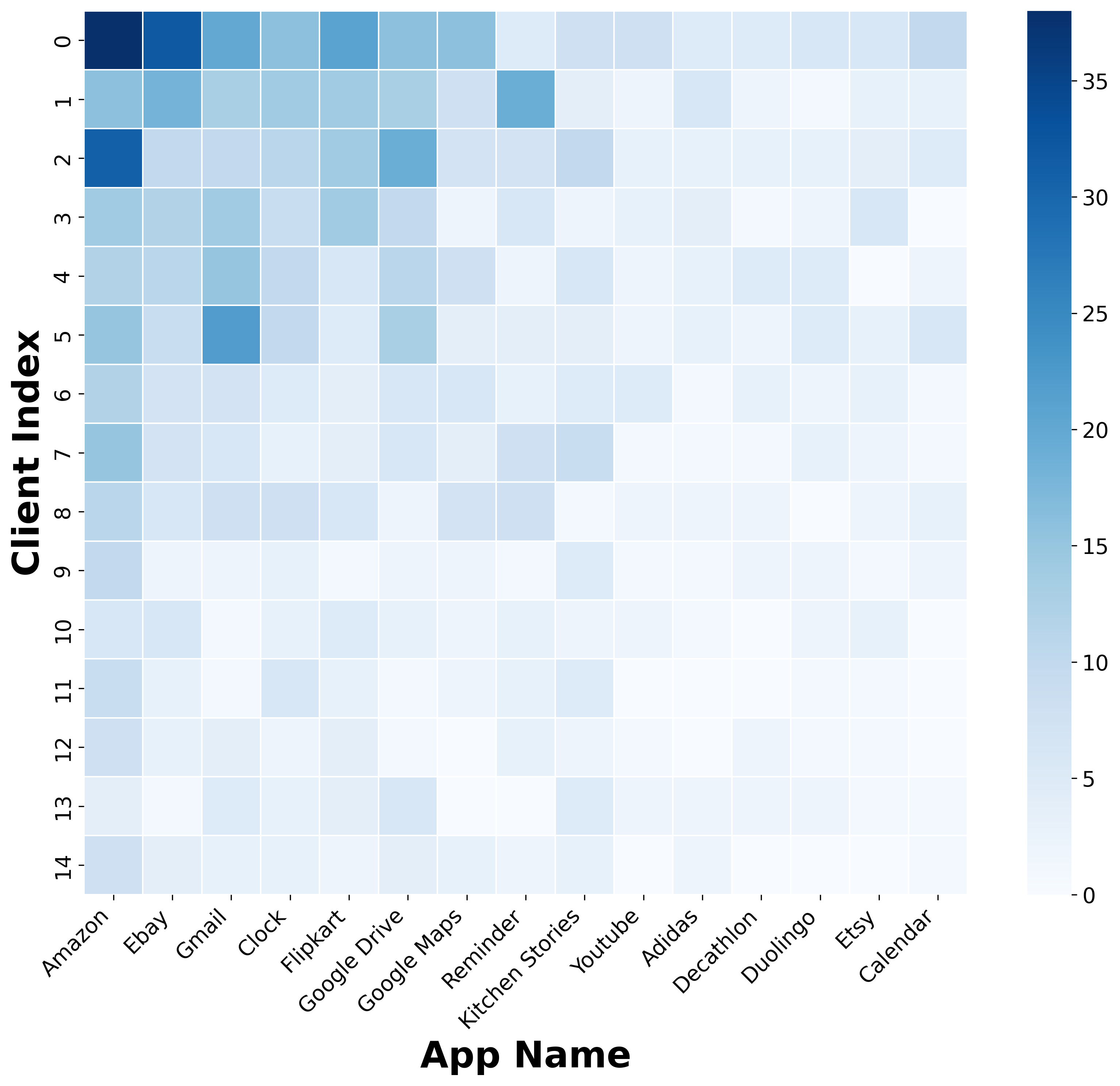}}
		\centerline{(c) ScaleApp Random}
	\end{minipage}
\caption{Heatmap distribution of the ScaleApp Dataset. We select top 15 apps for visualization.}
\label{fig:hetmap_ScaleApp}
\end{figure*}

\begin{acks}
This research is supported by the National Key R\&D Program of China under Grant 2021ZD0112801, NSFC under Grant 62171276 and the Science and Technology Commission of Shanghai Municipal under Grant 21511100900 and 22DZ2229005. 
\end{acks}

\section{Discussions and Future Directions}
\label{sec:discuss}
Previously, we have shown the promising results achieved by training mobile agents via federated learning. However, this is not the end as there are still emerging challenges and interesting directions that are worth exploring in this field in the future.

\subsection{Federated Algorithms for Heterogeneity of Mobile User Data}
\label{sec:discuss_heterogeneity}
In FedMABench, we establish the benchmark for evaluating federated mobile agents trained on heterogeneous user data. Our results in Section \ref{sec:exp_category} and \ref{sec:exp_app_level} demonstrate that currently no existing federated algorithm can achieve consistently good result when meeting the heterogeneity of diverse app usage. Specifically, FedYogi \cite{fedopt} has noticeable performance drop on the Category Half-Skew subset; FedMobileAgent \cite{wang2025fedmobileagenttrainingmobileagents} has no improvement when the distribution is completely skewed.

Recently, there has been some research \cite{mikaberidze2024consensus, yao2024ferrari, li2024filling} on generative AI and communication optimization for heterogeneous mobile clients. However, none of these studies address app heterogeneity among users.
The deployment of federated mobile agents require enhanced performance over diverse data distributions for scalability, which necessitates further research into designing novel FL algorithms to address the heterogeneity of phone usage trajectories.

\subsection{Privacy Preservation in Federated Mobile Agents}
\label{sec:discuss_privacy}

Training on user data inevitably raises privacy concerns.  
While federated learning helps mitigate privacy leakage by keeping private data on the client side and transmitting only LoRA adapters, potential privacy issues remain.

Models with substantial sizes are prone to memorization of their training data \cite{yuPrivacyPreservingInstructionsAligning2024, wang-etal-2024-knowledgesg}.  
Similar to large LLMs, recent studies \cite{caldarella2024phantommenaceunmaskingprivacy, samson2024privacyawarevisuallanguagemodels, jayaraman2024dejavumemorizationvisionlanguage} reveal that VLMs also inadvertently memorize and potentially expose sensitive information.  
Dejavu memorization \cite{jayaraman2024dejavumemorizationvisionlanguage} proposes a novel measurement for memorization by quantifying the fraction of ground-truth objects in an image that can be predicted from its text description in a training image-text pair.

Mobile agents rely on VLMs to perceive the interface and make decisions. Therefore, training directly on user data may lead to leakage of sensitive information.  
This issue can be addressed by implementing differential privacy (DP), which, however, remains underexplored in the context of VLMs and mobile agent training.

\begin{table*}[t]
\centering
\setlength\tabcolsep{5pt}
\caption{Experiments on the ScaleApp Dataset. Skewed app distribution results in lower average accuracy across apps. The long tailed apps with few episodes witness a greater decrease in performance. }
\vspace{-1mm}
\begin{tabular}{ll|cccccc}
\toprule
\textbf{Hetero.} & \textbf{Algorithm} & Amazon & Clock & Ebay & Flipkart & Gmail & \textbf{Avg.}  \\ 
\midrule
\rowcolor{gray!10} - & Zero-Shot & 30.68 & 32.53 & 39.84 & 33.33 & 18.35 & \cellcolor{gray!20} 32.17  \\ 
\rowcolor{gray!10} ~ & Central & 62.50 & 68.67 & 61.72 & 65.38 & 63.29 & \cellcolor{gray!20} 63.10  \\ 
\midrule

\rowcolor{red!10} ~ & Local 0 & 50.00 & 54.22 & 51.56 & 55.13 & 34.18 & \cellcolor{red!20} 46.72  \\ 
\rowcolor{red!10} ~ & Local 1 & 43.75 & 51.81 & 39.06 & 46.15 & 34.81 & \cellcolor{red!20} 44.83  \\ 
\rowcolor{red!10} ~ & FedAvg & 54.55 & 60.24 & 55.47 & 64.10 & 47.47 & \cellcolor{red!20} 54.35  \\ 
\rowcolor{red!10} ~ & FedProx & 54.55  & 59.04  & 55.47  & \textbf{65.38}  & 46.84  & \cellcolor{red!20} 54.46   \\ 
\rowcolor{red!10} ~ & FedAvgM & 54.55  & 59.04  & 55.47  & 61.54  & 46.84  & \cellcolor{red!20} 54.76   \\ 
\rowcolor{red!10} \multirow{-6}{*}{\cellcolor{red!10} ScaleApp IID} & FedYogi & \textbf{56.25}  & \textbf{61.45}  & \textbf{56.25}  & 64.10  & \textbf{48.10}  & \cellcolor{red!20} \textbf{55.12}  \\ 
\midrule

\rowcolor{red!10} ~ & Local 0 & 52.27 & 54.22 & 53.91 & 53.85 & 38.61 & \cellcolor{red!20} 49.67  \\ 
\rowcolor{red!10} ~ & Local 1 & 21.59 & 31.33 & 28.12 & 28.21 & 20.25 & \cellcolor{red!20} 25.61  \\ 
\rowcolor{red!10} ~ & FedAvg & 59.66 & \textbf{60.24} & 57.81 & 57.69 & 46.84 & \cellcolor{red!20} 55.35  \\ 
\rowcolor{red!10} ~ & FedProx & 58.52  & \textbf{60.24}  & 57.81  & 60.26  & \textbf{49.37}  & \cellcolor{red!20} 55.59   \\ 
\rowcolor{red!10} ~ & FedAvgM & \textbf{60.80}  & \textbf{60.24}  & \textbf{60.16}  & \textbf{62.82}  & 46.84  & \cellcolor{red!20} \textbf{55.71}   \\ 
\rowcolor{red!10} \multirow{-6}{*}{\makecell[{{l}}]{ScaleApp Random}} & FedYogi & 59.66  & 57.83  & 57.03  & 60.26  & 43.04  & \cellcolor{red!20} 53.93  \\ 
\midrule
\rowcolor{blue!10} & Local 0 & \textbf{59.66} & 39.76 & 54.69 & \textbf{61.54} & 26.58 & \cellcolor{blue!20} 42.46  \\ 
\rowcolor{blue!10}
~ & Local 1 & 48.30 & 49.40 & 48.44 & 52.56 & 28.48 & \cellcolor{blue!20} 41.87  \\ 
\rowcolor{blue!10}
~ & FedAvg & 57.39 & 57.83 & 55.47 & 58.97 & 40.51 & \cellcolor{blue!20} 52.81  \\ 
\rowcolor{blue!10}
~ & FedProx & 57.95  & 57.83  & 56.25  & 57.69  & 42.41  & \cellcolor{blue!20} 53.40   \\ 
\rowcolor{blue!10}
~ & FedAvgM & 57.95  & \textbf{59.04}  & \textbf{58.59}  & 60.26  & 43.04  & \cellcolor{blue!20} \textbf{54.41}   \\ 
\rowcolor{blue!10} \multirow{-6}{*}{ScaleApp Skew} & FedYogi & 58.52  & 57.83  & 54.69  & 60.26  & \textbf{43.67}  & \cellcolor{blue!20} 54.29  \\ 

\bottomrule
\end{tabular}
\label{tab:hetero_scaled_app}
\end{table*}

\subsection{Efficiency and Resources in Federated Mobile Agents}
\label{sec:discuss_efficiency}
To collaboratively train a global mobile agent on distributed user data, each user needs to locally train a small-sized VLM and communicate with the central server.  
However, limited computation resources and communication channels on mobile devices may hinder the feasibility of deployment.  

With the recent advancement of LLMs and diffusion models and their integration into federated learning systems \cite{zhou2021tea}, numerous approaches have been proposed to alleviate computational and communication overheads \cite{ding2024efedllm, raje2024communication, fang2025federatedsketchingloraondevice}.
On the other hand, the proliferation of smaller VLMs has significantly enhanced efficiency. For instance, AppVLM \cite{papoudakis2025appvlmlightweightvisionlanguage} specifically targets app control tasks with a lightweight architecture, facilitating rapid and cost-efficient inference for real-time execution.

\subsection{Combination of Reinforcement Learning with Federated Mobile Agents}
Although our current framework does not yet incorporate reinforcement learning, we identify it as a promising future direction. In a federated mobile agent setting, user feedback can serve as a critical reward signal, enabling agents to adjust their decision-making policies dynamically. 

Future work will need to tackle challenges inherent to integrating reinforcement learning into a federated environment, such as handling heterogeneous feedback, ensuring robust and stable learning under variable network conditions, and preserving user privacy. We believe that exploring these issues will pave the way for more adaptive and user-centric mobile agents, ultimately enhancing both their responsiveness and overall utility.

\section{Additional Experiments}
\subsection{Experiments on ScaleApp Dataset}
\textbf{Setups.}
We construct three subsets of the ScaleApp Dataset to further investigate the heterogeneity of specific app preferences.  
The distribution of subsets are visualized in the heatmaps in Figure \ref{fig:hetmap_ScaleApp}. We select the top 15 apps to plot as the rest 15 apps have basically the same distribution with the \(14\)-th app.
To enhance scalability and increase diversity, we select 30 apps, each with a varying number of episodes, to form a training set consisting of 2,500 episodes. Additionally, we sample 10\% of the episodes from each app to form the test set.

\textbf{Results.}
From Table \ref{tab:hetero_scaled_app}, we draw the following conclusions:
(1) By comparing FedAvg across the three subsets, we further confirm the presence of app-level heterogeneity, as a clear performance drop occurs when the model transitions to more heterogeneous scenarios.  
(2) Additionally, we observe that in heterogeneous settings, apps with a long-tailed distribution and fewer episodes experience a more significant performance decline compared to apps with more abundant data, such as Amazon and Ebay.  
(3) The performance of the \(0\)-th local client on Amazon in the ScaleApp Skew subset aligns with expectations, as the client has 300 training episodes of Amazon data. However, it also performs exceptionally well on Flipkart, even though it has not encountered any Flipkart data during training. This remarkable performance suggests that there may be shared patterns between Amazon and Flipkart, contributing to the unexpected yet correlated success.

\begin{table*}[t]
\centering
\setlength\tabcolsep{3pt}
\caption{Supplementary experiments on the Category-Level Dataset with more baselines. Colors represent \colorbox{red!10}{homogeneity} and \colorbox{blue!10}{heterogeneity}. FedMA is short for FedMobileAgent \cite{wang2025fedmobileagenttrainingmobileagents}. }
\vspace{-1mm}
\begin{tabular}{l|cccccc|l|cccccc}
\toprule
\textbf{Algorithm} & Shopping & Traveling & Office & Lives & Entertain. & \textbf{Avg.} & \textbf{Algorithm} & Shopping & Traveling & Office & Lives & Entertain. & \textbf{Avg.} \\
\midrule
Zero-Shot & \cellcolor{gray!10}26.61 & \cellcolor{gray!10}25.33 & \cellcolor{gray!10}27.05 & \cellcolor{gray!10}24.41 & \cellcolor{gray!10}23.81 & \cellcolor{gray!20}25.46 & Central & \cellcolor{gray!10}57.26 & \cellcolor{gray!10}58.67 & \cellcolor{gray!10}51.64 & \cellcolor{gray!10}55.12 & \cellcolor{gray!10}60.95 & \cellcolor{gray!20}56.90 \\ 
\midrule

\textbf{Homo.} & \multicolumn{6}{c|}{\cellcolor{red!10}\textbf{Category IID}} & \textbf{Hetero.} & \multicolumn{6}{c}{\cellcolor{blue!10}\textbf{Category Skew}}\\
Local 0 & \cellcolor{red!10}48.39 & \cellcolor{red!10}45.78 & \cellcolor{red!10}36.89 & \cellcolor{red!10}32.28 & \cellcolor{red!10}45.71 & \cellcolor{red!20}42.25 & Local 0 & \cellcolor{blue!10}50.81 & \cellcolor{blue!10}47.56 & \cellcolor{blue!10}46.72 & \cellcolor{blue!10}38.58 & \cellcolor{blue!10}48.57 & \cellcolor{blue!20}46.51 \\ 
FedAdagrad & \cellcolor{red!10}\textbf{54.84} & \cellcolor{red!10}\textbf{53.78} & \cellcolor{red!10}50.00 & \cellcolor{red!10}\textbf{39.37} & \cellcolor{red!10}\textbf{50.48} & \cellcolor{red!20}\textbf{50.21} & FedAdagrad & \cellcolor{blue!10}\textbf{54.03} & \cellcolor{blue!10}52.44 & \cellcolor{blue!10}\textbf{48.36} & \cellcolor{blue!10}\textbf{42.52} & \cellcolor{blue!10}\textbf{51.43} & \cellcolor{blue!20}\textbf{50.07} \\
SCAFFOLD & \cellcolor{red!10}53.23 & \cellcolor{red!10}52.00 & \cellcolor{red!10}\textbf{53.28} & \cellcolor{red!10}38.58 & \cellcolor{red!10}\textbf{50.48} & \cellcolor{red!20}49.79 & SCAFFOLD & \cellcolor{blue!10}\textbf{54.03} & \cellcolor{blue!10}\textbf{52.89} & \cellcolor{blue!10}47.54 & \cellcolor{blue!10}41.73 & \cellcolor{blue!10}\textbf{51.43} & \cellcolor{blue!20}49.93 \\

\midrule

\textbf{Hetero.} & \multicolumn{6}{c|}{\cellcolor{blue!10}\textbf{Category Half-Skew}} & \textbf{Hetero.} & \multicolumn{6}{c}{\cellcolor{blue!10}\textbf{Category Non-Uniform}}\\
Local 0 & \cellcolor{blue!10}41.13 & \cellcolor{blue!10}\textbf{56.00} & \cellcolor{blue!10}36.89 & \cellcolor{blue!10}\textbf{40.16} & \cellcolor{blue!10}37.14 & \cellcolor{blue!20}44.38 & Local 0 & \cellcolor{blue!10}38.71 & \cellcolor{blue!10}33.78 & \cellcolor{blue!10}34.43 & \cellcolor{blue!10}34.65 & \cellcolor{blue!10}33.33 & \cellcolor{blue!20}34.85 \\ 
FedAdagrad & \cellcolor{blue!10}\textbf{47.58} & \cellcolor{blue!10}46.22 & \cellcolor{blue!10}40.98 & \cellcolor{blue!10}35.43 & \cellcolor{blue!10}40.95 & \cellcolor{blue!20}42.82 & FedAdagrad & \cellcolor{blue!10}\textbf{50.00} & \cellcolor{blue!10}\textbf{52.89} & \cellcolor{blue!10}\textbf{49.18} & \cellcolor{blue!10}43.31 & \cellcolor{blue!10}\textbf{48.57} & \cellcolor{blue!20}\textbf{49.36} \\
SCAFFOLD & \cellcolor{blue!10}46.77 & \cellcolor{blue!10}48.89 & \cellcolor{blue!10}\textbf{42.62} & \cellcolor{blue!10}37.80 & \cellcolor{blue!10}\textbf{42.86} & \cellcolor{blue!20}\textbf{44.52} & SCAFFOLD & \cellcolor{blue!10}47.58 & \cellcolor{blue!10}52.00 & \cellcolor{blue!10}47.54 & \cellcolor{blue!10}\textbf{44.09} & \cellcolor{blue!10}\textbf{48.57} & \cellcolor{blue!20}48.51 \\

\bottomrule
\end{tabular}
\label{tab:hetero_category_supplement1}
\end{table*}

\begin{table*}[t]
\centering
\setlength\tabcolsep{2.7pt}
\caption{Supplementary experiments on the two other subsets of Category-Level Dataset: App Random and App Skew. Compared to the results in Category Skew, App Skew produces more severe heterogeneity. All FL algorithms demonstrate diverse performances on the two subsets with FedAvgM generally achieves the best results.}
\vspace{-1mm}
\begin{tabular}{l|cccccc|l|cccccc}
\toprule
\textbf{Algorithm} & Shopping & Traveling & Office & Lives & Entertain. & \textbf{Avg.} & \textbf{Algorithm} & Shopping & Traveling & Office & Lives & Entertain. & \textbf{Avg.} \\
\midrule
Zero-Shot & \cellcolor{gray!10}26.61 & \cellcolor{gray!10}25.33 & \cellcolor{gray!10}27.05 & \cellcolor{gray!10}24.41 & \cellcolor{gray!10}23.81 & \cellcolor{gray!20}25.46 & Central & \cellcolor{gray!10}57.26 & \cellcolor{gray!10}58.67 & \cellcolor{gray!10}51.64 & \cellcolor{gray!10}55.12 & \cellcolor{gray!10}60.95 & \cellcolor{gray!20}56.90 \\ 
\midrule

\textbf{Homo.} & \multicolumn{6}{c|}{\cellcolor{red!10}\textbf{App Random}} & \textbf{Hetero.} & \multicolumn{6}{c}{\cellcolor{blue!10}\textbf{App Skew}}\\

Local 0 & \cellcolor{red!10} 43.55 & \cellcolor{red!10} 45.78 & \cellcolor{red!10} 35.25 & \cellcolor{red!10} 43.31 & \cellcolor{red!10} 38.10 & \cellcolor{red!20} 41.96 & Local 0 & \cellcolor{blue!10} 44.35 & \cellcolor{blue!10} 40.44 & \cellcolor{blue!10} \textbf{48.36} & \cellcolor{blue!10} 29.92 & \cellcolor{blue!10} 35.24 & \cellcolor{blue!20} 39.83 \\
FedAvg & \cellcolor{red!10} 50.81 & \cellcolor{red!10} 51.56 & \cellcolor{red!10} 47.54 & \cellcolor{red!10} 44.09 & \cellcolor{red!10} 48.57 & \cellcolor{red!20} 48.93 & FedAvg & \cellcolor{blue!10} 50.81 & \cellcolor{blue!10} 53.78 & \cellcolor{blue!10} 45.90 & \cellcolor{blue!10} 33.86 & \cellcolor{blue!10} 53.33 & \cellcolor{blue!20} 48.22 \\
FedProx & \cellcolor{red!10} 49.19 & \cellcolor{red!10} 49.78 & \cellcolor{red!10} 46.72 & \cellcolor{red!10} 41.73 & \cellcolor{red!10} 49.52 & \cellcolor{red!20} 47.65 & FedProx & \cellcolor{blue!10} 51.61 & \cellcolor{blue!10} \textbf{54.22} & \cellcolor{blue!10} 47.54 & \cellcolor{blue!10} \textbf{38.58} & \cellcolor{blue!10} \textbf{54.29} & \cellcolor{blue!20} \textbf{49.79} \\
FedAvgM & \cellcolor{red!10} 50.00 & \cellcolor{red!10} \textbf{54.67} & \cellcolor{red!10} 46.72 & \cellcolor{red!10} 44.09 & \cellcolor{red!10} \textbf{52.38} & \cellcolor{red!20} \textbf{50.21} & FedAvgM & \cellcolor{blue!10} \textbf{52.42} & \cellcolor{blue!10} 52.00 & \cellcolor{blue!10} 45.90 & \cellcolor{blue!10} 37.01 & \cellcolor{blue!10} \textbf{54.29} & \cellcolor{blue!20} 48.65 \\
FedYogi & \cellcolor{red!10} \textbf{53.23} & \cellcolor{red!10} 51.56 & \cellcolor{red!10} \textbf{49.18} & \cellcolor{red!10} \textbf{46.46} & \cellcolor{red!10} 48.57 & \cellcolor{red!20} 50.07 & FedYogi & \cellcolor{blue!10} 50.00 & \cellcolor{blue!10} 52.89 & \cellcolor{blue!10} 45.08 & \cellcolor{blue!10} 35.43 & \cellcolor{blue!10} 49.52 & \cellcolor{blue!20} 47.37 \\

\bottomrule
\end{tabular}
\label{tab:hetero_category_supplement2}
\end{table*}

\subsection{Supplementary Experiments on Category-Level and App-Level Datasets}
\label{sec:exp_category_level_supplement}
\textbf{Setups.}
The experimental settings as the same with the experiments in Section \ref{sec:exp_category} and \ref{sec:exp_app_level}.
Due to page limits, we present more results with different baselines and other subsets in this section for reference.
We use "FedMA" to denote FedMobileAgent for spacing. The colors represent homogeneity and heterogeneity.

\textbf{Results.}
We draw the following conclusions:  
(1) As shown in Table \ref{tab:hetero_category_supplement1}, we further substantiate that training mobile agents using federated learning yields promising enhancements, as all baselines exhibit remarkable progress compared to local training.
(2) From Tables \ref{tab:hetero_category_supplement1} and \ref{tab:hetero_category}, global aggregation methods based on optimization (FedAdam, FedAdagrad, and FedYogi) consistently manifest subpar performance on the Category Half-Skew subset, but demonstrate exceptional results on the other subsets. This performance discrepancy remains challenging to explain.  
(3) By comparing the FL results on the two subsets, Category Skew and App Skew, in Tables \ref{tab:hetero_category_supplement2} and \ref{tab:hetero_category}, we conclude that FL algorithms generally underperform on the App Skew subset, which indicates that app name heterogeneity is more fundamental and severe than app category heterogeneity.
(4) As shown in Tables \ref{tab:app_level_supplement} and \ref{tab:app_level}, the eight baselines exhibit diverse performance across different heterogeneous scenarios. FedMobileAgent performs averagely, as it is not specifically designed to handle this type of heterogeneity, and it degrades to standard FedAvg when the app distribution becomes extremely skewed.  
(5) As reaffirmed, no current FL algorithm effectively addresses the new heterogeneity introduced by federated mobile agents, as all FL algorithms experience a substantial decline from IID to non-IID app distributions, which highlights the need for further advancements in this area.

\begin{table*}[t]
\centering
\setlength\tabcolsep{4.1pt}
\caption{Supplementary Experiments on the App-Level Dataset. We provide additional evaluation results with four other baselines. The total eight baselines yield diverse performance in different heterogeneous scenarios.}
\vspace{-1mm}
\begin{tabular}{l|cccccc|l|cccccc}
\toprule
\textbf{Algorithm} & Amazon & Clock & Ebay & Flipkart & Gmail & \textbf{Avg.} & \textbf{Algorithm} & Amazon & Clock & Ebay & Flipkart & Gmail & \textbf{Avg.} \\ 
\midrule
Zero-Shot & \cellcolor{gray!10}29.75 & \cellcolor{gray!10}32.38 & \cellcolor{gray!10}28.33 & \cellcolor{gray!10}30.00 & \cellcolor{gray!10}28.12 & \cellcolor{gray!20}29.62 & Central & \cellcolor{gray!10}54.55 & \cellcolor{gray!10}64.76 & \cellcolor{gray!10}58.33 & \cellcolor{gray!10}61.00 & \cellcolor{gray!10}51.56 & \cellcolor{gray!20}57.67 \\
\midrule
\textbf{Homo.} & \multicolumn{6}{c|}{\cellcolor{red!10}\textbf{App IID}} & \textbf{Hetero.} & \multicolumn{6}{c}{\cellcolor{blue!10}\textbf{App Skew}}\\
Local 0 & \cellcolor{red!10}44.63 & \cellcolor{red!10}49.52 & \cellcolor{red!10}41.67 & \cellcolor{red!10}50.00 & \cellcolor{red!10}33.59 & \cellcolor{red!20} \cellcolor{red!20}43.38 & Local 0 & \cellcolor{blue!10}\textbf{56.20} & \cellcolor{blue!10}36.19 & \cellcolor{blue!10}42.50 & \cellcolor{blue!10}44.00 & \cellcolor{blue!10}21.09 & \cellcolor{blue!20}39.72 \\
FedAdagrad & \cellcolor{red!10} 56.20 & \cellcolor{red!10} \textbf{54.29} & \cellcolor{red!10} 54.17 & \cellcolor{red!10} \textbf{58.00} & \cellcolor{red!10} \textbf{50.00} & \cellcolor{red!20} \textbf{54.36} & FedAdagrad & \cellcolor{blue!10} 45.45 & \cellcolor{blue!10} \textbf{54.29} & \cellcolor{blue!10} 50.83 & \cellcolor{blue!10} \textbf{55.00} & \cellcolor{blue!10} \textbf{46.88} & \cellcolor{blue!20} \textbf{50.17} \\ 
SCAFFOLD & \cellcolor{red!10} 56.20 & \cellcolor{red!10} \textbf{54.29} & \cellcolor{red!10} \textbf{55.00} & \cellcolor{red!10} 53.00 & \cellcolor{red!10} 47.66 & \cellcolor{red!20} 53.14 & SCAFFOLD & \cellcolor{blue!10} 48.76 & \cellcolor{blue!10} \textbf{54.29} & \cellcolor{blue!10} \textbf{52.50} & \cellcolor{blue!10} 52.00 & \cellcolor{blue!10} 44.53 & \cellcolor{blue!20} \textbf{50.17} \\ 
FedMA & \cellcolor{red!10} \textbf{58.68} & \cellcolor{red!10} 53.33 & \cellcolor{red!10} 53.33 & \cellcolor{red!10} 55.00 & \cellcolor{red!10} 48.44 & \cellcolor{red!20} 53.66 & FedMA & \cellcolor{blue!10} 47.93 & \cellcolor{blue!10} 53.33 & \cellcolor{blue!10} 48.33 & \cellcolor{blue!10} 54.00 & \cellcolor{blue!10} 41.41 & \cellcolor{blue!20} 48.61 \\ 

\midrule
\textbf{Hetero.} & \multicolumn{6}{c|}{\cellcolor{blue!10}\textbf{App Half-Skew}} & \textbf{Hetero.} & \multicolumn{6}{c}{\cellcolor{blue!10}\textbf{App Non-Uniform}}  \\
Local 0 & \cellcolor{blue!10}52.89 & \cellcolor{blue!10}\textbf{57.14} & \cellcolor{blue!10}\textbf{45.00} & \cellcolor{blue!10}40.00 & \cellcolor{blue!10}36.72 & \cellcolor{blue!20}46.17 & Local 0 & \cellcolor{blue!10}39.67 & \cellcolor{blue!10}\textbf{58.10} & \cellcolor{blue!10}38.33 & \cellcolor{blue!10}48.00 & \cellcolor{blue!10}\textbf{46.09} & \cellcolor{blue!20}45.64 \\ 

FedAdagrad & \cellcolor{blue!10} 54.55 & \cellcolor{blue!10} 54.29 & \cellcolor{blue!10} 43.33 & \cellcolor{blue!10} \textbf{55.00} & \cellcolor{blue!10} \textbf{42.19} & \cellcolor{blue!20} 49.48 & FedAdagrad & \cellcolor{blue!10} \textbf{56.20} & \cellcolor{blue!10} 54.29 & \cellcolor{blue!10} \textbf{46.67} & \cellcolor{blue!10} 50.00 & \cellcolor{blue!10} 40.62 & \cellcolor{blue!20} 49.30 \\
SCAFFOLD & \cellcolor{blue!10} 54.55 & \cellcolor{blue!10} 53.33 & \cellcolor{blue!10} 43.33 & \cellcolor{blue!10} 54.00 & \cellcolor{blue!10} 40.62 & \cellcolor{blue!20} 48.78 & SCAFFOLD & \cellcolor{blue!10} 55.37 & \cellcolor{blue!10} 53.33 & \cellcolor{blue!10} 45.83 & \cellcolor{blue!10} 50.00 & \cellcolor{blue!10} 40.62 & \cellcolor{blue!20} 48.78 \\ 
FedMA & \cellcolor{blue!10} \textbf{55.37} & \cellcolor{blue!10} 53.33 & \cellcolor{blue!10} 45.83 & \cellcolor{blue!10} 55.00 & \cellcolor{blue!10} 41.41 & \cellcolor{blue!20} \textbf{49.83} & FedMA & \cellcolor{blue!10} 55.37 & \cellcolor{blue!10} 55.24 & \cellcolor{blue!10} 45.83 & \cellcolor{blue!10} \textbf{52.00} & \cellcolor{blue!10} 41.41 & \cellcolor{blue!20} \textbf{49.65} \\ 

\bottomrule
\end{tabular}
\label{tab:app_level_supplement}
\vspace{-1mm}
\end{table*}

\subsection{Comparison of Base Models}
\label{sec:exp_base_model}

\textbf{Setups.}  
Built upon ms-swift, FedMABench supports over ten base VLMs and has the potential to accommodate more in the future.  
We select five models as representatives, encompassing both open-ended and closed-ended models from three distinct model families.  
Since closed-ended models cannot be fine-tuned, we provide zero-shot results for them.  
For open-ended models, we fine-tune them on the App IID subset of the App-Level Dataset as a representative case.

\textbf{Results.}  
As shown in Table \ref{tab:ablation_base_model}, we draw the following conclusions:  
(1) Training on different models yields diverse performance results.  
(2) Overall, the performance of open-ended models shows a strong positive correlation with their model size.  
(3) Through federated training on distributed data, even smaller VLMs like Qwen2-VL-2B-Instruct can achieve performance on par with SOTA closed-ended models such as GPT-4o.

\begin{table}[t]
\centering
\setlength\tabcolsep{4.1pt}
\caption{Comparison of different base models on the App IID subset. We choose five models as representatives including both open-ended and closed-ended models.}
\vspace{-1mm}
\begin{tabular}{l|cccccc}
\toprule
\textbf{Base Model} & Amazon & Clock & Ebay & Flipkart & Gmail & \textbf{Avg.} \\
\midrule
\textbf{Algorithm} & \multicolumn{6}{c}{\cellcolor{gray!10}\textbf{Zero-Shot}}  \\ 
GPT-4o & \cellcolor{gray!10} 40.50 & \cellcolor{gray!10} 48.57 & \cellcolor{gray!10} 43.33 & \cellcolor{gray!10} 45.00 & \cellcolor{gray!10} 38.28 & \cellcolor{gray!20} 42.86 \\ 
GPT-4o-mini & \cellcolor{gray!10} 26.45 & \cellcolor{gray!10} 33.33 & \cellcolor{gray!10} 30.83 & \cellcolor{gray!10} 30.00 & \cellcolor{gray!10} 35.16 & \cellcolor{gray!20} 31.18 \\ 
\midrule
\textbf{Algorithm} & \multicolumn{6}{c}{\cellcolor{red!10}\textbf{FedAvg}}  \\ 
Qwen2-VL-2B & \cellcolor{red!10} 47.11 & \cellcolor{red!10} 46.67 & \cellcolor{red!10} 35.83 & \cellcolor{red!10} 38.00 & \cellcolor{red!10} 39.84 & \cellcolor{red!20} 41.46 \\ 
Qwen2-VL-7B & \cellcolor{red!10} 57.02 & \cellcolor{red!10} 53.33 & \cellcolor{red!10} 52.50 & \cellcolor{red!10} 55.00 & \cellcolor{red!10} 46.88 & \cellcolor{red!20} 52.79 \\ 
InternVL2-1B & \cellcolor{red!10} 28.93 & \cellcolor{red!10} 40.00 & \cellcolor{red!10} 27.50 & \cellcolor{red!10} 28.00 & \cellcolor{red!10} 35.16 & \cellcolor{red!20} 31.88 \\ 
InternVL2-2B & \cellcolor{red!10} 34.71 & \cellcolor{red!10} 41.90 & \cellcolor{red!10} 30.00 & \cellcolor{red!10} 28.00 & \cellcolor{red!10} 32.03 & \cellcolor{red!20} 33.28 \\ 
\bottomrule
\end{tabular}
\label{tab:ablation_base_model}
\vspace{-2mm}
\end{table}

\subsection{Ablation on Dataset Size}
\label{sec:exp_data_size}
\textbf{Setups.}
We conduct experiments on the Basic-AC Dataset with incrementally increasing data sizes to investigate the impact of dataset size on performance, and to examine whether scaling laws hold in the context of federated learning for mobile agent training. 
To control experimental conditions, we fix the number of clients at 10 and evaluate the mobile agents after 10 communication rounds. Notably, in the FedAvg implementation, 30\% of participating clients are randomly sampled per round, leading to a smaller number of sample iterations compared to centralized training.

\textbf{Results.}
As shown in Table \ref{tab:ablation_dataset}, we draw the following conclusions:
(1) Performance improvements exhibit a strong positive correlation with dataset scale across all training paradigms, validating the effectiveness of federated learning for scalable mobile agent training. Specifically, FedAvg demonstrates incremental gains from 31.18\% to 53.54\% as data availability increases.
(2) FedAvg shows diminishing returns as the data size reaches a certain threshold, still leaving a gap relative to centralized training. Enhancing the performance of federated trained mobile agents necessitates further efforts into this area.

\subsection{Ablation on Clients Number}
\label{sec:exp_client_number}
\textbf{Setups.}
We investigate federated learning dynamics under varying client number while maintaining a fixed budget of 100 episodes per client. Mobile agents are evaluated after 100 training rounds with a controlled participation scheme: each round activates 10\% of available clients.

\textbf{Results.}
As shown in Table \ref{tab:ablation_cnumber}, we conclude that:
(1) As reiterated, model performance demonstrates strong positive correlation with client population size, validating federated learning's effectiveness for scalable distributed training.
(2) A particularly significant performance leap (51.81\% → 56.06\% step accuracy) occurs when scaling from 10 to 30 clients, suggesting critical mass benefits in collaborative learning.

\subsection{Ablation on Clients Participation}
\label{sec:exp_client_partition}
\textbf{Setups.}
We analyze the impact of client participation rates while keeping the total client population constant and maintaining a fixed global data volume. Specifically, we use the subset of Basic-AC with 3,000 episodes, partitioned across 30 clients.
The system is evaluated after 100 training rounds with varying numbers of clients sampled per round, ranging from 1 to 30 participants.

\textbf{Results.}
As shown in Table \ref{tab:ablation_cpar}, we draw the following conclusions:
(1) Cross-referencing with Table \ref{tab:ablation_cnumber} reveals an emergent pattern: under equivalent total data budgets, increasing client participation enhances model performance. This suggests distributed learning benefits stem not merely from data accumulation, but crucially from diversified experiential sampling across heterogeneous clients.
(2) Moderate participation rates, with 3 clients sampled per round, achieve performance comparable to maximum participation. This phenomenon can be attributed to the fact that as the number of participating clients increases, heterogeneity also rises, which may degrade overall performance despite the higher training cost.



\begin{table}[t]
\centering
\setlength\tabcolsep{5pt}
\caption{Experiments on dataset sizes. Performance improvements exhibit strong positive correlation with dataset scale for all training paradigms.}
\begin{tabular}{l|cccccc}
\toprule
\textbf{Algorithm} & 200 & 500 & 1000 & 3000 & 5000 & 7000  \\ 
\midrule
Zero-Shot & \multicolumn{6}{c}{27.24} \\ 
Central & 43.94 & 42.36 & 55.59 & 56.38 & 59.69 & 62.05 \\ 
\midrule
Local 0 & 17.80 & 28.35 & 37.64 & 44.25 & 47.40 & 52.44 \\ 
FedAvg & 31.18 & 36.54 & 43.78 & 50.39 & 51.50 & 53.54 \\ 

\bottomrule
\end{tabular}
\label{tab:ablation_dataset}
\end{table}

\begin{table}[t]
\centering
\setlength\tabcolsep{5pt}
\caption{
Experiments with different client numbers. Each client is allocated 100 episodes. As more clients are involved, the dataset scale increases. Performance improvements show a positive correlation with the number of clients, consistent with the results in Table \ref{tab:ablation_dataset}.
}
\begin{tabular}{l|ccccc}
\toprule
        Client Number  & 10 & 30 & 50 & 70 \\ 
        Client Sample  & 1 & 3 & 5 & 7 \\ 
        \midrule
        FedAvg& 51.81 & 56.06 & 57.17 & 57.48 \\ 
\bottomrule
\end{tabular}
\label{tab:ablation_cnumber}
\end{table}

\begin{table}[t]
\centering
\setlength\tabcolsep{5pt}
\caption{
Experiments with varying client participation rates, with the dataset and its partition kept constant for controlled comparison. A moderate number of clients per round achieves comparable performance to full participation.}
\begin{tabular}{l|cccccc}
\toprule
        
        Client Number & 30 & 30 & 30 & 30 & 30 & 30 \\ 
        Client Sample & 1 & 3 & 5 & 10 & 15 & 30 \\ 
        \midrule
        FedAvg & 41.10 & 45.35 & 44.72 & 44.09 & 43.94 & 45.67 \\ 

\bottomrule
\end{tabular}
\label{tab:ablation_cpar}
\end{table}

\section{Data \& Experiment Details}
\label{app:ExperimentalDetails}
\subsection{Dataset Details}
\label{app:dataset_details}
We provide detailed descriptions of our datasets and the data collection process in this section, including examples and statistics.

\begin{figure}
    \centering
    \includegraphics[width=1\linewidth]{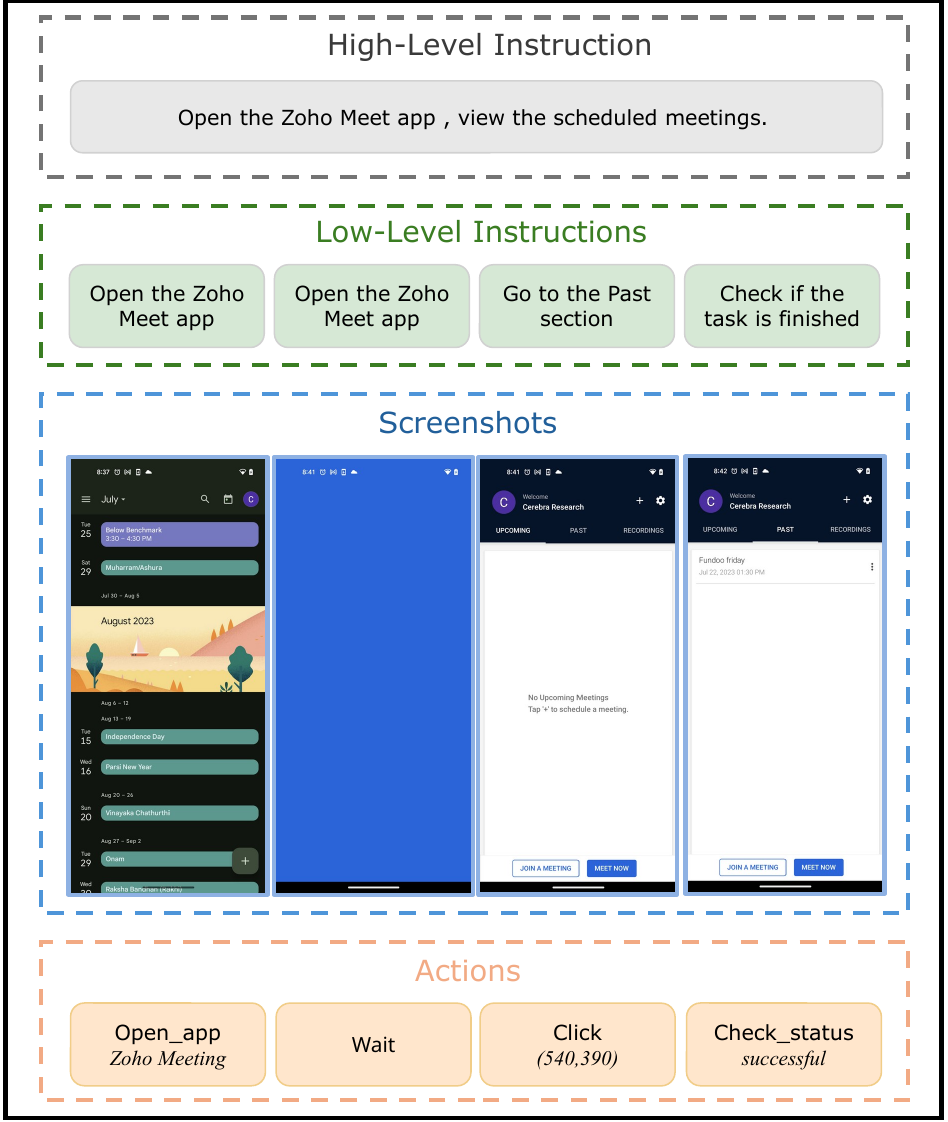}
    \vspace{-5mm}
    \caption{A data episode example for training mobile agents.}
    \label{fig:example}
    \vspace{-2mm}
\end{figure}

\textbf{Data Episode Example.}
To provide a clearer understanding of the structure of our dataset and the composition of a data episode, we present a sample as an example in this section. 
As shown in Figure \ref{fig:example}, each episode consists of: 
(1) A high-level instruction, which is a natural language sentence describing the task to be accomplished;
(2) A sequence of low-level instructions, detailing the fine-grained tasks required for the current screenshot;
(3) A series of screenshots taken from the start to the end of the task; and
(4) A corresponding list of actions, matching the number of screenshots, indicating what the user does to progress to the next screenshot. 
All actions belong to an action space containing 7-9 options. We adopt the action spaces defined in \cite{rawlesAndroidWildLargeScale2023, liEffectsDataScale2024, wang2025fedmobileagenttrainingmobileagents}.

\textbf{Dual-Strategy App Name Extraction.}
For each episode from the original dataset of Android Control, we implement a dual-strategy approach for application name extraction based on the "open app" action and regular expression matching.

As demonstrated in the following code snippet, if the actions include the the "open app" action, the application name is directly retrieved from the dedicated \texttt{app\_name} field, followed by sanitized string processing.

\begin{lstlisting}
for episode in all_episodes:
    // Load the task information for the current episode
    data = load("task_info.json")
    // Check if 'open_app' in the actions of this episode
    if "open_app" in data["action_type"]:
        // Direct extraction with sanitization
        app_name = data["app_name"].replace("\ufeff", "") 
    else:
        // Use semantic pattern matching for extraction from the "goal" instruction
        pattern = re.compile(r'\bthe\s+(\w+(?:\s+\w+)?)\s+app\b', re.IGNORECASE)
        match = pattern.search(data["goal"])
        if match exists:
            app_name = match.group(1)
        else:
            // Skip episodes where extraction fails
            continue
\end{lstlisting}
Otherwise, if the actions of this episode do not contain the "open app" action, which indicates that explicit application identifiers are absent, we attempt to extract potential app name from the \texttt{goal} field (i.e., instructions) . This is achieved through a regular expression designed to identify the phrase \texttt{"the [app] app"} using semantic pattern matching.

Episodes failing both extraction strategies were systematically excluded to ensure data validity. This dual-strategy filtering process ultimately yielded 8,400 qualified episodes containing unambiguous application identifiers, forming the core dataset for subsequent construction and analysis.

\begin{table*}[t]
\centering
\setlength\tabcolsep{6pt}
\caption{
Composition details of the 14 subsets in the Basic-AC Dataset. 
'N.' denotes the number of instances, and 'Category' refers to the covered categories within each subset.
}
\label{tab:Basic-ACSpecifics}
\begin{tabular}{l|ccccc}
\toprule
        Dataset & Subset & Category & N. Client & N. Episode & N. Step\\ 
        \midrule
        \multirow{14}{*}{Basic-AC} & c10n200 & all & 10 & 200 & 7454\\ 
        &c10n500 & All & 10 & 500 & 20198\\ 
        &c10n1000 & All & 10 & 1000 & 40112 \\ 
        &c10n3000 & All & 10 & 3000 & 120512\\ 
        &c10n5000 & All & 10 & 5000 & 201434\\ 
        &c10n7000 & All & 10 & 7000 & 282332\\ 
        &c30n3000 & All & 30 & 3000 & 120512 \\ 
        &c50n5000 & All & 50 & 5000 & 201434\\ 
        &c70n7000 & All & 70 & 7000 & 282332\\ 
        &Shopping & Shopping & 10 & 2252 & 79292\\ 
        &Travelling & Travelling & 10 & 788 & 47918\\ 
        &Office & Office & 10 & 1974 & 76910\\ 
        &Lives & Lives & 10 & 1136 & 46070\\ 
        &Entertainment & Entertainment & 10 & 850 & 32150\\ 
    
\bottomrule
\end{tabular}

\end{table*}

\textbf{Dataset Statistics.}
In this part, we provide a detailed enumeration of the specific apps included in each dataset, along with the exact number of instances for each app.

(1) Basic-AC Dataset:
The Basic-AC Dataset encompasses comprehensive categories and apps. Detailed statistical information can be found in Table \ref{tab:apps_categorization}.
(2) Category-Level Dataset:
The Basic Dataset comprises a total of 52 apps that are organized into several categories. 
In the shopping category, there are 10 apps: Amazon, eBay, Flipkart, Adidas, Nike, Decathlon, Etsy, Puma, Temu, and Snapdeal, with each app contributing 20 instances for a total of 200. 
The travelling category includes 10 apps, namely Google Maps, Expedia, Omio, Booking.com, Citymapper, Trainline, Kayak, Cruisemapper, MakeMyTrip, and Agoda, where each app again provides 20 instances to reach a sum of 200.
The office category follows the same pattern with 10 apps: Gmail, Clock, Google Drive, Google Docs, Calendar, Google Keep, Contacts, Reminder, Recorder, and Voice Recorder, each adding 20 data points for a total of 200. 
The lives category also consists of 10 apps: Kitchen Stories, Home Workout, Sidechef, Yummly, Blossom, Plantum, Simple Habit, Leafsnap, Medito, and Insight Timer, each contributing 20 instances to make up another 200. 
In contrast, the entertainment category is slightly different, comprising 12 apps. Eight of these apps, which are YouTube, Vimeo, Artsy, Sketchbook, Messenger, Pinterest, Flipboard, and SoundCloud, each provide 20 instances, while the remaining four apps, namely Snapchat, SmartNews, The Hindu, and CNN, contribute 10 instances each, together totaling 200.

\textbf{Basic-AC Specifics.}
We construct 14 subsets in Basic-AC, a detailed description of which is provided in Table \ref{tab:Basic-ACSpecifics}. The table specifies three key parameters for each subset: number of participating clients, total episodes, and total steps. Subsets 1-9 represent cross-category aggregations with varying scales, while subsets 10-14 correspond to category-specific partitions.

\subsection{Training Details}
\textbf{General Parameters.}
Our implementation leverages the Swift library \cite{zhao2024swiftascalablelightweightinfrastructure} with parameter-efficient fine-tuning. The LoRA configuration employs a rank of 8 with an alpha scaling factor of 32, incorporating dropout regularization of 0.05 to prevent overfitting. We set the maximum sequence length to 4,096. 
We set the batch size to 1 and the gradient accumulation step to 4. The learning rate is kept fixed at 5e-5.

\textbf{Hardware Configuration.}
The training is conducted on two NVIDIA GeForce RTX 3090 GPUs utilizing CUDA version 12.4. Under this hardware configuration, the training process achieves a throughput of approximately 2 minutes per training round per client when processing 10 episodes. 

\textbf{Federated Algorithms.}  
The framework implements adaptive hyperparameter defaults for various federated algorithms: FedYogi \cite{fedopt} employs momentum factors $(\beta_1=0.9, \beta_2=0.999)$ with learning rate $\eta=10^{-3}$ and stabilization constant $\tau=10^{-6}$. FedAvgM \cite{fedavgm} uses $0.9/0.1$ ratio for historical/current model interpolation. FedProx \cite{fedprox} applies proximal regularization with $\mu=0.2$ through $||w-w^t||^2$ penalty terms. SCAFFOLD \cite{scaffold} configurations maintain server learning rate $\eta_s=1.0$ with client momentum compensation, while FedAdam and FedAdagrad \cite{fedopt} share base parameters $(\beta_1=0.9, \beta_2=0.999)$ with adaptive learning rate scaling. All algorithms expose tunable coefficients through the framework's unified parameter interface.

\clearpage
\onecolumn

\begin{center}    
\small 
\begin{longtable}[ht]{p{1.6cm}r|p{1.6cm}r|p{1.6cm}r|p{1.6cm}r|p{1.6cm}r|p{1.6cm}r}

\caption{Application categorization and statistics for the Basic-AC Dataset. Due to the limited table width, app names that are too long will be truncated, with the truncated portion replaced by a dot(.).} \label{tab:apps_categorization} \\

\toprule
\textbf{App} & \textbf{Num} & \textbf{App} & \textbf{Num} & \textbf{App} & \textbf{Num} & \textbf{App} & \textbf{Num} & \textbf{App} & \textbf{Num} & \textbf{App} & \textbf{Num} \\
\midrule
\endfirsthead
\toprule
\textbf{App} & \textbf{Num} & \textbf{App} & \textbf{Num} & \textbf{App} & \textbf{Num} & \textbf{App} & \textbf{Num} & \textbf{App} & \textbf{Num} & \textbf{App} & \textbf{Num} \\
\midrule
\endhead
			
\hline 
\multicolumn{3}{l}{{Continued on next page.}} \\ 
\endfoot

\hline
\endlastfoot

\multicolumn{12}{l}{\textbf{Shopping}}\\
amazon & 302 & ebay & 225 & flipkart & 151 & adidas & 83 & decathlon & 82 & etsy & 76 \\
nike & 64 & temu & 64 & puma & 59 & shopsy & 53 & snapdeal & 52 & ikea & 47 \\
shopclues & 43 & ubuy & 40 & banggood & 38 & industrybuy. & 37 & myntra & 37 & tata cliq & 37 \\
zara & 37 & jiomart & 33 & dhgate & 32 & blinkit & 30 & moglix & 30 & bigbasket & 26 \\
asos & 25 & joom & 20 & tata neu & 20 & dmart ready & 19 & ajio & 17 & hardware sh. & 17 \\
nnnow & 17 & pepperfry & 17 & edmunds & 15 & houzz & 13 & footshop & 12 & hamleys & 12 \\
limeroad & 12 & rapidbox & 12 & mywarehouse & 11 & nykaaman & 11 & toys 'r' us & 11 & yalla toys & 11 \\
coolblue & 10 & freshtohome & 10 & lazada & 10 & mega hardwa. & 10 & shoppers st. & 10 & barakat & 9 \\
cartrade & 8 & furlenco & 8 & nykaafashion & 8 & autoscout24 & 7 & lovelocal & 7 & cars24 & 6 \\
carwale & 6 & nykaa & 6 & olx india & 6 & spinny & 6 & toyspoint & 6 & woodenstreet & 6 \\
dookanti & 5 & uniqlo & 5 & urbanic & 5 & albertsons & 4 & hardware sh. & 4 & jd & 4 \\
louis vuitt. & 4 & max fashion & 4 & nature's ba. & 4 & pdffiller & 4 & sports dire. & 4 & true value & 4 \\
urban outfi. & 4 & zalando & 4 & 1800 flowers & 3 & abercrombie & 3 & adani one & 3 & bechdo & 3 \\
bewakoof & 3 & carguru & 3 & dunzo & 3 & globalsourc. & 3 & homzmart & 3 & igp & 3 \\
khelmart & 3 & nykaa fashi. & 3 & peter engla. & 3 & pizza max & 3 & reliance di. & 3 & shoptime & 3 \\
spencers & 3 & sportsuncle & 3 & westside & 3 & cardekho & 2 & colourpop c. & 2 & coop & 2 \\
ferns n pet. & 2 & flower aura & 2 & funeasylearn & 2 & furniture o. & 2 & instashop & 2 & jaquar & 2 \\
louis phili. & 2 & love local & 2 & m\&s india & 2 & magzter & 2 & massimo dut. & 2 & milkbasket & 2 \\
moira cosme. & 2 & namshi & 2 & noon & 2 & p louise co. & 2 & pantaloons & 2 & pepper & 2 \\
redbubble & 2 & royal & 2 & safeway & 2 & sports bazar & 2 & sportsdirect & 2 & sportspar & 2 \\
super note & 2 & top{-}most ha. & 2 & topmost har. & 2 & weather rad. & 2 & yoox & 2 & zappo & 2 \\
zappo brands & 2 & acme & 1 & apkpure & 1 & app market & 1 & character c. & 1 & dubizzle & 1 \\
ebay app & 1 & electronics. & 1 & estee lauder & 1 & farfetch & 1 & fernsnpetals & 1 & goat & 1 \\
gostor & 1 & ikea  app & 1 & industry ub. & 1 & insaraf {-} s. & 1 & iplan.ai & 1 & jd sports & 1 \\
jollee & 1 & kicks crew & 1 & luxuryestate & 1 & massimo  du. & 1 & mikbasket & 1 & mytrip & 1 \\
nnnnow & 1 & nobroker & 1 & same temu & 1 & samsung shop & 1 & sanitary ba. & 1 & second cale. & 1 \\
sun \& sand . & 1 & tesco & 1 & thriftbooks & 1 & toys shoppi. & 1 & tradet mark. & 1 & vijetha live & 1 \\
winni & 1 & woodland & 1 & woodlands & 1 & zomato & 1 &  &  &  &  \\
\hline

\multicolumn{12}{l}{\textbf{Travelling}}\\
google maps & 111 & expedia & 55 & omio & 47 & booking.com & 46 & kayak & 40 & citymapper & 37 \\
cruisemapper & 29 & makemytrip & 27 & trainline & 27 & airbnb & 25 & skyscanner & 25 & agoda & 23 \\
wanderu & 21 & alltrails & 20 & rail planner & 20 & guardian & 15 & moovit & 14 & traillink & 13 \\
hopper & 11 & momondo & 11 & rome2rio & 11 & trip.com & 11 & yatra & 10 & cruisedeals & 9 \\
goibibo & 9 & amtrak & 8 & easemytrip & 8 & ixigo & 8 & klook & 8 & flixbus & 7 \\
foursquare & 7 & talabat & 7 & time zone c. & 6 & trainpal & 6 & schedule pl. & 5 & cleartrip & 4 \\
kiwi.com & 4 & shipatlas & 4 & traveloka & 4 & getby & 3 & hiking proj. & 3 & hotels.com & 3 \\
immobiliare & 3 & lambus & 3 & maxmilhas & 3 & prestigia & 3 & rail europe & 3 & riyadh bus & 3 \\
travel life & 3 & wego flight. & 3 & bookaway & 2 & eurostar & 2 & gotogate & 2 & greyhound & 2 \\
hhr train & 2 & hiiker & 2 & klm & 2 & lner & 2 & orbitz & 2 & passporter & 2 \\
sbb mobile & 2 & sncf connect & 2 & sygic travel & 2 & trovit & 2 & cheapflights & 1 & egy train w. & 1 \\
farefirst & 1 & maps go & 1 & mytrip & 1 & roadtrippers & 1 & sncb intern. & 1 & thalys & 1 \\
trivago & 1 &  &  &  &  &  &  &  &  &  &  \\
\hline

\multicolumn{12}{l}{\textbf{Office}}\\
gmail & 189 & clock & 158 & google drive & 127 & reminder & 101 & calendar & 72 & contacts & 69 \\
google keep & 65 & google docs & 52 & recorder & 48 & voice recor. & 47 & google slid. & 45 & ticktick & 37 \\
khan aca. & 36 & skype & 36 & chat & 33 & powerpoint & 33 & settings & 31 & files by go. & 30 \\
dropbox & 28 & officesuite & 22 & todoist & 22 & phonebook & 21 & polaris off. & 20 & clockbuddy & 19 \\
all currenc. & 18 & memrise & 17 & microsoft w. & 17 & onedrive & 17 & outlook & 17 & smart recor. & 16 \\
google news & 15 & taskito & 15 & tasks & 15 & jotform & 14 & myrecorder & 14 & any.do & 13 \\
readera & 13 & translate & 13 & currency pl. & 12 & easy voice . & 12 & migros & 12 & merriam. & 11 \\
to do remin. & 11 & to do list & 10 & formsapp & 9 & notein & 9 & presentatio. & 9 & colornote & 8 \\
coursera & 8 & easy dialer & 8 & easy notes & 8 & easy timezo. & 8 & xodo & 8 & zoho meeting & 8 \\
calculator & 7 & note & 7 & spck editor & 7 & webex & 7 & alarmy & 6 & dictionary & 6 \\
duocards & 6 & habitica & 6 & meet & 6 & microsoft p. & 6 & mondly lang. & 6 & moon+ reader & 6 \\
pdf reader . & 6 & whiteboard & 6 & notebook & 5 & pcloud & 5 & schedule pl. & 5 & simple calc. & 5 \\
timezone co. & 5 & alarm clock. & 4 & calendar pl. & 4 & code editor & 4 & digital ala. & 4 & easynotes & 4 \\
forms app & 4 & plantapp & 4 & savvy time & 4 & sheets & 4 & sublime text & 4 & tododo & 4 \\
vocab.com & 4 & webex meet & 4 & winzip & 4 & word office & 4 & zarchiver & 4 & alarm clock. & 3 \\
clevnote & 3 & contact & 3 & cursa & 3 & cx file exp. & 3 & deftpdf & 3 & digical & 3 \\
doodle & 3 & forms.app & 3 & math tests & 3 & my money & 3 & pdfelement & 3 & pull\&bear & 3 \\
spendee & 3 & udemy & 3 & voice recor. & 3 & weather xl & 3 & calcu & 2 & calendar pro & 2 \\
carrot & 2 & ereader pre. & 2 & flipsnack & 2 & funeasylearn & 2 & giant stopw. & 2 & google tasks & 2 \\
letter temp. & 2 & math learni. & 2 & microsoft 3. & 2 & multi calcu. & 2 & munimobile & 2 & mycurrency & 2 \\
papago & 2 & power point & 2 & pro 7{-}zip & 2 & quip & 2 & simple cont. & 2 & smartcal & 2 \\
super note & 2 & timezones & 2 & unit conver. & 2 & voice recor. & 2 & world clock & 2 & xe converter & 2 \\
zoho show & 2 & 7z & 1 & blaze wordp. & 1 & bookscape & 1 & calculator . & 1 & currencycon. & 1 \\
docx {-} all . & 1 & drawing pad & 1 & everand ebo. & 1 & exchange ra. & 1 & focus to{-}do & 1 & g{-}forms & 1 \\
internal fi. & 1 & iplan.ai & 1 & lists & 1 & math learni. & 1 & maths test & 1 & monefy & 1 \\
office: pre. & 1 & oppia & 1 & pdf extra & 1 & radio u.s. & 1 & rar & 1 & setting & 1 \\
simple clock & 1 & smartify & 1 & step tracke. & 1 & telegram & 1 & upgrad & 1 & webcode & 1 \\
\hline

\multicolumn{12}{l}{\textbf{Lives}}\\
kitchen sto. & 91 & home work. & 51 & fit & 50 & sidechef & 44 & yummly & 44 & insight tim. & 38 \\
leafsnap & 30 & redfin & 30 & blossom & 28 & weather & 27 & plantum & 26 & opentable & 25 \\
google fit & 24 & simple habit & 24 & plantin & 23 & strava & 20 & dmart ready & 19 & fitai & 19 \\
fitbit & 19 & meditopia & 19 & idanim & 18 & artier & 17 & medito & 17 & pepperfry & 17 \\
calm & 15 & jefit & 15 & grubhub & 13 & mindfulness & 13 & tasty & 13 & cookpad & 12 \\
deliveroo & 12 & evolve & 12 & migros & 12 & trovit homes & 12 & breethe & 11 & lifestyle & 11 \\
photos & 11 & supercook & 11 & all recipes & 10 & bigoven & 10 & lunch recip. & 10 & notes & 10 \\
doordash & 9 & home centre & 9 & 99acres & 8 & all recipes. & 8 & rentberry & 8 & urban ladder & 8 \\
fitpro & 7 & heartfulness & 7 & phases of t. & 7 & talabat & 7 & bbc news & 6 & budgetbytes & 6 \\
daff moon & 6 & moon & 6 & pizza hut & 6 & withings & 6 & baby tracker & 5 & balance & 5 \\
gym workout & 5 & martinoz pi. & 5 & mi fitness & 5 & my moon pha. & 5 & plant ident. & 5 & realtor.com & 5 \\
runkeeper & 5 & housing & 4 & moonx & 4 & serenity & 4 & flo & 3 & headspace & 3 \\
healthifyme & 3 & ovia pregna. & 3 & planta & 3 & pregnancy & 3 & smiling mind & 3 & trulia & 3 \\
carrot & 2 & hatch baby & 2 & home garden & 2 & immoscout24 & 2 & moonly & 2 & plantora & 2 \\
property fi. & 2 & recime & 2 & vivareal & 2 & what to exp. & 2 & babycenter & 1 & cult.fit & 1 \\
freshto home & 1 & good food & 1 & immobiliare. & 1 & indian reci. & 1 & luxuryestate & 1 & mojopizza & 1 \\
my workout . & 1 & nobroker & 1 & workout pla. & 1 &  &  &  &  &  &  \\
\hline

\multicolumn{12}{l}{\textbf{Entertainment}}\\
youtube & 95 & vimeo & 64 & gallery & 36 & artsy & 35 & messenger & 32 & pinterest & 31 \\
spotify & 27 & sketchbook & 26 & soundcloud & 23 & flipboard & 20 & snapchat & 17 & the weather. & 16 \\
cnn & 15 & google news & 15 & guardian & 15 & arts \& cult. & 13 & tunein radio & 13 & wynk music & 12 \\
audiomack & 11 & deviantart & 11 & nytimes & 11 & photos & 11 & pocketbook & 11 & show & 11 \\
smartnews & 11 & youtube mus. & 11 & coolblue & 10 & mytuner rad. & 10 & the hindu & 10 & sgraffito & 9 \\
skyview free & 9 & behance & 8 & reuters & 7 & sketchar & 7 & bbc news & 6 & moon+ reader & 6 \\
radio garden & 6 & time zone c. & 6 & toi & 6 & washington . & 6 & webnovel & 6 & whiteboard & 6 \\
color & 5 & euronews & 5 & gaana & 5 & hindu & 5 & kobo books & 5 & mi fitness & 5 \\
thefork & 5 & wattpad & 5 & cafeyn & 4 & cna & 4 & cnn news & 4 & dolby on & 4 \\
domino's & 4 & hindu news & 4 & hungama & 4 & usa today & 4 & anghami & 3 & dailymotion & 3 \\
fox news & 3 & headspace & 3 & mojarto & 3 & nbc news & 3 & peggy & 3 & rtistiq & 3 \\
toi news & 3 & daily art & 2 & dailyart & 2 & hiiker & 2 & magzter & 2 & msn weather & 2 \\
paint & 2 & radio & 2 & radio fm & 2 & readly & 2 & readwhere m. & 2 & sky tracker & 2 \\
startracker & 2 & zinio magaz. & 2 & app market & 1 & artly & 1 & bbdaily & 1 & deccan hera. & 1 \\
expert pape. & 1 & hipaint & 1 & messages & 1 & newyork tim. & 1 & radio u.s. & 1 & readly maga. & 1 \\
sky view & 1 & skyview & 1 & smartify & 1 & winni & 1 &  &  &  &  \\
\bottomrule



\end{longtable}
\end{center}

\clearpage
\twocolumn

\end{document}